%% file: main.tex
\documentclass[11pt,a4paper]{article}
\usepackage[T1]{fontenc}
\usepackage[utf8]{inputenc}
\usepackage[margin=1in]{geometry}
\usepackage{amsmath,amssymb,amsfonts,amsthm}
\usepackage{algorithm,algorithmic}
\usepackage{booktabs,bm,bbm,tabularx,threeparttable}
\usepackage{longtable,array,multirow,enumitem}
\usepackage{graphicx,xcolor,textcomp,verbatim,placeins}
\usepackage[round,authoryear]{natbib}
\usepackage{xurl}
\usepackage[hidelinks]{hyperref}
\usepackage[capitalise]{cleveref}
\newtheorem{theorem}{Theorem}
\hypersetup{pdftitle={Geophysics-informed invertible operator network for Bayesian geophysical inversion},
pdfauthor={Yangwen Zhang, Xintong Wang, Xiaofei Guan, Peng Yu}}
\title{Geophysics-informed invertible operator network for Bayesian geophysical inversion}
\author{\parbox{0.96\textwidth}{\centering
Yangwen Zhang$^{1}$, Xintong Wang$^{2}$, Xiaofei Guan$^{1,*}$, Peng Yu$^{3}$\\[0.8em]
\small
$^{1}$School of Mathematical Sciences, Tongji University,\\
1239 Siping Road, Shanghai 200092, China\\[0.4em]
$^{2}$Department of Mathematics, Shanghai Normal University,\\
100 Guilin Road, Shanghai 200234, China\\[0.4em]
$^{3}$State Key Laboratory of Marine Geology, School of Ocean and Earth Science,\\
Tongji University, 1239 Siping Road, Shanghai 200092, China\\[0.8em]
\footnotesize
\href{mailto:2311747@tongji.edu.cn}{2311747@tongji.edu.cn};
\href{mailto:xtwang@shnu.edu.cn}{xtwang@shnu.edu.cn};\\
\href{mailto:guanxf@tongji.edu.cn}{guanxf@tongji.edu.cn};
\href{mailto:yupeng@tongji.edu.cn}{yupeng@tongji.edu.cn}\\[0.3em]
$^{*}$Corresponding author: Xiaofei Guan}}
\date{}
\begin{document}
\maketitle
\begin{abstract}
Geophysical inverse problems are inherently ill-posed because of sparse observations and measurement noise. Classical sampling-based Bayesian inference methods, such as Markov chain Monte Carlo (MCMC), are computationally prohibitive because they require many forward simulations. Data-driven machine-learning methods often struggle to produce physically consistent solutions when training data are limited. To address these issues, we propose a geophysics-informed invertible operator network (GI-ION) that integrates an invertible neural network (INN) within a deep operator-learning framework for efficient Bayesian geophysical inversion. The INN serves as the branch network and transforms the model parameters into branch coefficients that represent the forward solution and a latent variable that captures uncertainty in the inverse problem. By learning the forward operator and enforcing statistical independence between the branch coefficients and the latent variable, GI-ION enables efficient sampling from a learned conditional approximation, even with sparse boundary observations. We validated GI-ION on three representative cases: 1D surface-wave dispersion and travel-time tomography for a 2D synthetic random-field model and a seismic-slice model. Numerical experiments demonstrate robust performance across diverse problem settings and efficient conditional sampling.
\end{abstract}

\section{Introduction}\label{sec:introduction}
Geophysical inversion aims to infer subsurface properties from surface or subsurface observations and plays an important role in resource exploration, geoinformation mapping, and geohazard monitoring \cite{gosselin2022review, linde2015geological}. Because the observations are typically indirect, incomplete, and noisy, geophysical inverse problems are often ill-posed and require reliable and efficient inversion frameworks.

Traditional geophysical inversion methods generally use regularization or Bayesian inference. Regularization methods, including Tikhonov approaches, stabilize the inversion by imposing explicit constraints on the solution space. However, they usually return a single deterministic estimate and therefore provide limited uncertainty information \cite{golub1999tikhonov, rodriguez2013total, calvetti2018inverse}. Bayesian methods characterize posterior uncertainty more naturally, but often become computationally expensive because likelihood evaluation requires repeated forward simulations \cite{sen1996bayesian, grana2017bayesian, deng2023efficient}. This cost has motivated accelerated and surrogate approaches \cite{nawaz2019rapid, conrad2016accelerating, frangos2010surrogate}. These limitations motivate more efficient probabilistic inversion frameworks.

In recent years, deep learning techniques have attracted considerable attention for geophysical inverse problems \cite{yu2021deep}. These methods are typically data-driven, learning nonlinear mappings from observations to subsurface model parameters from labeled datasets. For example, convolutional neural networks (CNNs) have been successfully applied to electromagnetic inversion \cite{puzyrev2019deep, oh2018salt, puzyrev2021inversion}. More recent studies have also incorporated probabilistic inference into deep models. Bayesian convolutional neural networks, for instance, have been introduced for MT inversion to provide both resistivity predictions and uncertainty estimates \cite{poddar2025inversion}, while implicit neural representations combined with variational inference have been used for large-scale seismic inversion \cite{romero2025bayesian}. Transformer-based frameworks have further extended multimodal geophysical inversion by jointly incorporating different data types within a unified architecture equipped with uncertainty estimation \cite{jiang2025one}. In parallel, reduced-order and representation-learning approaches have also been explored for probabilistic inversion, including singular-value-based basis learning and compressed latent representations \cite{corrales2024seis2rock, rincon2025integrating, manassero2020reduced}. Despite these advances, conventional deep learning approaches remain largely data-driven, often requiring extensive labeled datasets and providing solutions with limited physical consistency when observations are sparse or noisy \cite{schuster2024review, yu2024learning}. This limitation motivates the integration of physical constraints into the learning process.

The physics-informed paradigm provides a principled way to embed governing laws into the learning process, thereby improving physical consistency and generalization. Physics-informed neural networks (PINNs) are a representative example of this idea, combining physical residuals and data misfits within a unified objective \cite{karniadakis2021physics}. In geophysical inversion, PINNs have been applied to seismic full waveform inversion, electrical resistivity tomography, and seismic tomography, demonstrating that physically constrained learning can reduce the reliance on labeled ground-truth models \cite{rasht2022physics, waheed2021pinntomo, baldan2023physics}. However, PINNs are typically tailored to a single problem instance, so changes in the physical setting often require retraining and may lead to substantial computational cost \cite{okazaki2022physics}.

Recent advances in operator learning have sought to overcome this limitation by learning mappings between function spaces. This approach reduces prediction cost across multiple problem instances. Representative frameworks include Fourier and convolutional neural operators and DeepONet-type architectures. These frameworks can represent solution operators on both regular and irregular domains \cite{lifourier, raonic2023convolutional, li2020multipole, tranfactorized}. Further examples and extensions are provided by \cite{li2023geometry, lu2021learning, leehyperdeeponet}. To address data scarcity, physics-informed extensions such as PI-DeepONet and PINO have been developed \cite{li2024physics, wang2021learning}. In geophysical inverse problems, neural operators have shown promise in seismic imaging, full waveform inversion, and electromagnetic inversion, offering both computational efficiency and robustness to noisy or incomplete data \cite{zong2024randomized, zong2024solving}.

Despite these advances, the ill-posedness of geophysical inverse problems still makes it difficult to learn a stable and unique mapping from observations to model parameters. Small perturbations in noisy observations may lead to large variations in the inferred solution, motivating a shift from point estimation to posterior approximation. Within this Bayesian perspective, invertible neural networks (INNs) provide a natural mechanism for density estimation and efficient posterior sampling \cite{ardizzoneanalyzing, winkler2019learning, radev2020bayesflow}. Recent studies have further combined invertible architectures with physics-informed constraints, such as invertible DeepONet, and introduced latent-noise mechanisms to represent aleatoric uncertainty \cite{kaltenbach2023semi, zhang2021bayesian}. However, reliable posterior approximation under simultaneously sparse observations and limited training data remains insufficiently explored in geophysical settings. This gap motivates the GI-ION framework proposed in this work.

In this work, we introduce the geophysics-informed invertible operator network (GI-ION), a probabilistic operator-learning framework tailored to Bayesian geophysical inversion.
The term ``geophysics-informed'' reflects the intrinsic challenges of geophysical inverse problems, in which subsurface properties must often be inferred from sparse, indirect, and noisy observations. In such settings, purely data-driven operator learning may reproduce the available measurements but still yield physically inconsistent solutions in unobserved regions, especially when observational coverage is limited. GI-ION addresses this issue by incorporating the governing geophysical equations into the training objective through a physics-informed loss function. Rather than relying solely on pointwise data supervision, this loss regularizes the learned operator over the full computational domain and promotes global consistency with the underlying geophysical laws.
A second key innovation lies in GI-ION's unique operator architecture, which is based on DeepONet. We replace the standard deterministic branch network with an invertible neural network (INN). This design decomposes the input model into two statistically independent components: (i) branch coefficients, which encode the deterministic forward physics within a learned basis space, and (ii) a latent noise variable, which captures the inherent uncertainties of the problem.
This fully invertible architecture enables GI-ION to perform efficient bidirectional inference, simultaneously learning the forward operator and constructing a conditional posterior approximation without iterative online sampling.
Furthermore, to improve conditional uncertainty estimation under limited data, we introduce an independence-oriented training objective. The conventional Maximum Mean Discrepancy (MMD) loss used in INNs exhibited stronger residual dependence in the experiments considered here. In contrast, the proposed objective directly penalizes dependence between the learned physical coefficients and the latent variable. This combination of a physics-embedded architecture and an independence-oriented constraint supports approximate conditional sampling in the considered data-scarce settings.

The remainder of this paper is organized as follows. Section~\ref{sec:problem} outlines the Bayesian geophysical inverse problem. Section~\ref{sec:methods} presents inverse operator learning and the GI-ION framework, including its training algorithm and inversion procedure. Section~\ref{sec:numerical-experiments} evaluates the proposed model using 1D surface-wave dispersion, 2D synthetic random-field tomography, and 2D seismic-slice tomography. Finally, Section~\ref{sec:conclusions} summarizes the conclusions and outlines directions for future research.

\section{Problem statement}\label{sec:problem}

Let $\boldsymbol{m}\in\mathcal{M}$ denote the subsurface model and let $\mathcal{F}:\mathcal{M}\rightarrow\mathcal{D}$ denote the noise-free geophysical forward operator. The observation model is
\begin{equation}
  \boldsymbol{d}
  \;=\;
  \mathcal{F}(\boldsymbol{m})+\boldsymbol{\varepsilon},
  \label{eq:forward-model}
\end{equation}
where $\boldsymbol{\varepsilon}$ denotes observational noise with density $p_{\varepsilon}$, $\boldsymbol{m}\sim p_0(\boldsymbol{m})$, and $p_0$ is the prior density. Under this additive-noise model, the likelihood is
\[
p(\boldsymbol{d}\mid\boldsymbol{m})
=
p_{\varepsilon}\!\left(\boldsymbol{d}-\mathcal{F}(\boldsymbol{m})\right).
\]
Bayes' rule gives the posterior
\begin{equation}
\label{eq:bayesian-posterior}
p(\boldsymbol{m}\mid\boldsymbol{d})
=
\frac{p(\boldsymbol{d}\mid\boldsymbol{m})p_0(\boldsymbol{m})}
{\displaystyle\int p(\boldsymbol{d}\mid\boldsymbol{m}')p_0(\boldsymbol{m}')\,\mathrm{d}\boldsymbol{m}'}.
\end{equation}
Because the inverse mapping is generally non-unique and unstable, we do not assume that a deterministic inverse $\mathcal{F}^{-1}$ exists. The objective of GI-ION is to construct a learned conditional distribution $q_\theta(\boldsymbol{m}\mid\boldsymbol{d})$ that approximates the Bayesian posterior,
\[
q_\theta(\boldsymbol{m}\mid\boldsymbol{d})
\approx
p(\boldsymbol{m}\mid\boldsymbol{d}),
\]
while permitting efficient conditional sampling after offline training.

\section{Methods}\label{sec:methods}
\subsection{Inverse operator learning}
\label{sec:inverse-operator}

Invertible neural networks (INNs) provide a bijective mapping between variables of equal total dimension \cite{dinh2014nice, dinh2017density, kingma2018glow}. Let $\boldsymbol{m}\in\mathbb{R}^{d_m}$ denote the vectorized model parameters. In GI-ION, instead of directly mapping $\boldsymbol{m}$ to observations, we define an invertible branch map
\begin{equation}
\label{eq:branch_map}
\mathcal{B}_{\theta}:\mathbb{R}^{d_m}\rightarrow \mathbb{R}^{P}\times \mathbb{R}^{d_z},
\qquad
\boldsymbol{m}\mapsto \bigl(\boldsymbol{b}(\boldsymbol{m}),\boldsymbol{z}(\boldsymbol{m})\bigr),
\end{equation}
where $\boldsymbol{b}(\boldsymbol{m})\in\mathbb{R}^{P}$ denotes the branch coefficients associated with the forward field, and $\boldsymbol{z}(\boldsymbol{m})\in\mathbb{R}^{d_z}$ is a latent variable introduced to capture the non-uniqueness of the inverse problem. Since $\mathcal{B}_{\theta}$ is invertible, the dimensions satisfy
\begin{equation}
\label{eq:dimension_balance}
d_m=P+d_z.
\end{equation}

The forward field is represented by a DeepONet-style operator,
\begin{equation}
\label{eq:forward_operator_factorized}
\mathcal{F}_{\theta}(\boldsymbol{m})(\boldsymbol{x})
=
\boldsymbol{b}(\boldsymbol{m})^{\top}\mathcal{T}_{\theta}(\boldsymbol{x}),
\end{equation}
where $\mathcal{T}_{\theta}(\boldsymbol{x})\in\mathbb{R}^{P}$ is the trunk output at query point $\boldsymbol{x}$. Under this formulation, the inverse reconstruction is performed through the inverse branch map
\begin{equation}
\label{eq:inverse_branch_map}
\boldsymbol{m}=\mathcal{B}_{\theta}^{-1}(\boldsymbol{b},\boldsymbol{z}).
\end{equation}

For a given observation vector $\boldsymbol{d}\in\mathbb{R}^{M_{\mathrm{o}}}$ sampled at observation locations $\boldsymbol{x}_{\mathrm{o}}=\{x_{\mathrm{o}}^1,\dots,x_{\mathrm{o}}^{M_{\mathrm{o}}}\}$, the coefficient vector is estimated by projecting the data onto the learned trunk basis:
\begin{equation}
\label{eq:observation_matrix}
\mathbf{T}_{\mathrm{o}}
=
\begin{bmatrix}
\mathcal{T}_{\theta}(x_{\mathrm{o}}^1)^{\top}\\
\vdots\\
\mathcal{T}_{\theta}(x_{\mathrm{o}}^{M_{\mathrm{o}}})^{\top}
\end{bmatrix},
\qquad
\boldsymbol{b}^{*}(\boldsymbol{d})
=
\mathbf{T}_{\mathrm{o}}^{\dagger}\boldsymbol{d},
\end{equation}
where $\dagger$ denotes the Moore--Penrose inverse. If $\mathbf{T}_{\mathrm{o}}$ has full column rank, $\boldsymbol{b}^{*}(\boldsymbol{d})$ is the unique least-squares estimate. If full column rank is not satisfied, the Moore--Penrose inverse returns the unique minimum-norm least-squares estimate. Conditional samples are generated by
\begin{equation}
\label{eq:posterior_sampling_rule}
\boldsymbol{m}
=
\mathcal{B}_{\theta}^{-1}\bigl(\boldsymbol{b}^{*}(\boldsymbol{d}),\boldsymbol{z}\bigr),
\qquad
\boldsymbol{z}\sim p(\boldsymbol{z})=\mathcal{N}(\mathbf{0},\mathbf{I}).
\end{equation}

GI-ION imposes independence between the coefficient vector and the latent variable in the learned representation:
\begin{equation}
\label{eq:factorized_joint_intro}
p_{\theta}(\boldsymbol{b},\boldsymbol{z})
=
p_{\theta}(\boldsymbol{b})\,p(\boldsymbol{z}).
\end{equation}
Under this condition, fixing $\boldsymbol{b}=\boldsymbol{b}^{*}(\boldsymbol{d})$ leaves the latent variable distributed according to its prescribed reference distribution:
\begin{equation}
\label{eq:conditional_z_is_standard_normal}
p_{\theta}\!\left(\boldsymbol{z}\mid \boldsymbol{b}=\boldsymbol{b}^{*}(\boldsymbol{d})\right)
=
p(\boldsymbol{z})
=
\mathcal{N}(\mathbf{0},\mathbf{I}).
\end{equation}
Define the conditional generation map
\[
g_{\theta,\boldsymbol{d}}(\boldsymbol{z})
:=
\mathcal{B}_{\theta}^{-1}\!\left(\boldsymbol{b}^{*}(\boldsymbol{d}),\boldsymbol{z}\right).
\]
The learned conditional distribution induced by GI-ION is the pushforward
\begin{equation}
\label{eq:conditional-pushforward}
q_{\theta}(\cdot\mid\boldsymbol{d})
:=
\left(g_{\theta,\boldsymbol{d}}\right)_{\#}p(\boldsymbol{z}),
\end{equation}
where $\#$ denotes the pushforward of a measure. Equivalently, samples are obtained by drawing $\boldsymbol{z}\sim p(\boldsymbol{z})$ and evaluating $\boldsymbol{m}=g_{\theta,\boldsymbol{d}}(\boldsymbol{z})$. The distribution $q_{\theta}(\cdot\mid\boldsymbol{d})$ is a learned conditional approximation to the Bayesian posterior rather than an exact closed-form posterior density. The overall framework and inference workflow of GI-ION are illustrated in \cref{fig:gi-ion-architecture}. In practice, standard MMD-based regularization may become unstable in high-dimensional geophysical settings; therefore, GI-ION further introduces physics-informed training and an independence loss to improve robustness and posterior approximation quality.

\subsection{Geophysics-informed invertible operator network}
\label{subsec:SemiSupervisedPINN}

\begin{figure*}[!t]
	\centering
	\includegraphics[width=1.0\textwidth]{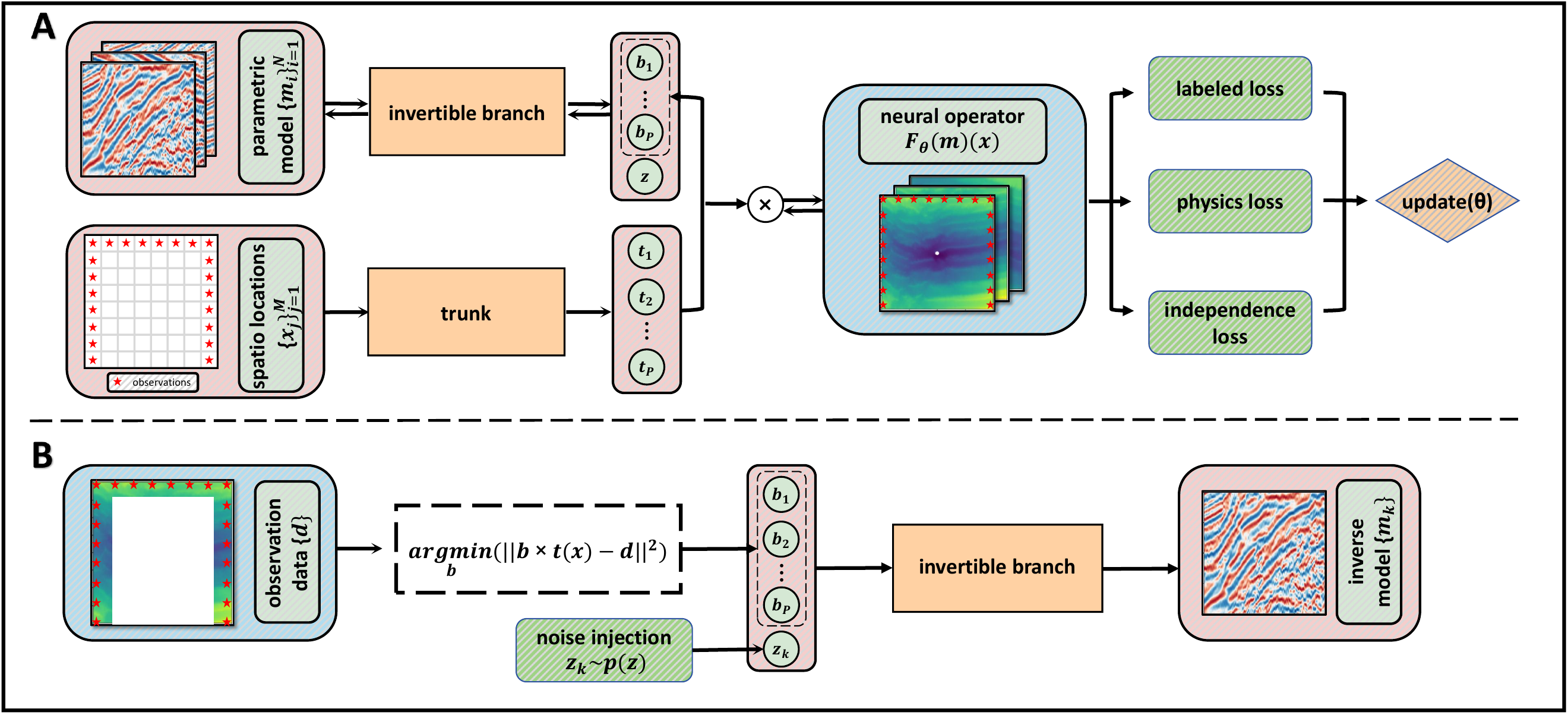}
	\caption{GI-ION schematic: (a) framework and training process and (b) inversion workflow after training. During inverse-operator evaluation, $\boldsymbol{z}$ is sampled from a predefined noise distribution $p(\boldsymbol{z})$.}
	 \label{fig:gi-ion-architecture}
\end{figure*}

As illustrated in \cref{fig:gi-ion-architecture}, GI-ION combines an invertible branch network with a trunk network to learn a physics-informed forward operator. The trunk network $\mathcal{T}_\theta(\boldsymbol{x})\in\mathbb{R}^{P}$ takes the query coordinate $\boldsymbol{x}$ as input and represents the basis functions of the output field, while the branch network $\mathcal{B}_\theta$ maps the model parameter $\boldsymbol{m}$ to the coefficient--latent pair $(\boldsymbol{b}(\boldsymbol{m}),\boldsymbol{z}(\boldsymbol{m}))$. The forward field is therefore expressed as
\begin{equation}
\begin{aligned}
	\label{eq:deeponet-expansion}
	\mathcal{F}_{\theta}(\boldsymbol{m})(\boldsymbol{x}) 
    &= \underbrace{\boldsymbol{b}(\boldsymbol{m})^{\top}}_{\text{branch}}  \cdot \underbrace{\mathcal{T}_\theta(\boldsymbol{x})}_{\text{trunk}}\\
    &= \sum_{j=1}^{P} b_j(\boldsymbol{m})  \cdot t_j(\boldsymbol{x}), 
\end{aligned}
\end{equation}
where $t_j(\boldsymbol{x})$ and $b_j(\boldsymbol{m})$ denote the $j$th components of the trunk basis and the branch coefficient vector, respectively. By construction, $\boldsymbol{b}(\boldsymbol{m})\in\mathbb{R}^{P}$ and $\mathcal{T}_{\theta}(\boldsymbol{x})\in\mathbb{R}^{P}$, while the latent variable satisfies $\boldsymbol{z}(\boldsymbol{m})\in\mathbb{R}^{d_z}$ and $d_m=P+d_z$.
This architecture comprises a branch network that maps model parameters to coefficient--latent representations and a trunk network that maps spatial coordinates to output-field basis functions.
The trunk network can receive different spatial coordinates to generate predictions for the corresponding observation positions. For problems involving variable observation sets, our framework accommodates this by assigning a corresponding set of query points to each sample.
Furthermore, the GI-ION architecture can be extended to different problem dimensions by adjusting the dimension of its basis-function space.

For GI-ION training, consider a model dataset $\{\boldsymbol{m}_i \}_{i=1}^{N}$ containing $N$ samples and a corresponding observation dataset $\left\{\boldsymbol{d}_i \, \middle|\, \left(d_i^1, \dots, d_i^{M_\text{o}} \right)^{\top} \right\}_{i=1}^{N}$. Each model has $M_\text{o}$ observation locations $\boldsymbol{x}_\text{o} = \left( x^1_\text{o}, \dots, x_\text{o}^{M_\text{o}}\right)^{\top}$ in \(\mathcal{D}\), together with unobserved query points $\boldsymbol{x}_\text{u}$.
The vector $\boldsymbol{x}_\text{total} = \left[\boldsymbol{x}_\text{o}, \boldsymbol{x}_\text{u}\right]^{\top}$ contains all query points, including observation locations $\boldsymbol{x}_\text{o}$ and unobserved points $\boldsymbol{x}_\text{u}$.
GI-ION uses a training strategy that combines data and physics-informed losses. In practice, $\mathcal{L}_{\text{physics}}$ is evaluated at query points distributed over the entire domain to promote global physical consistency, whereas $\mathcal{L}_{\text{data}}$ enforces agreement between the forward solution and the observation data.
The physical term is derived from the governing equation of the corresponding physical process, which provides a physical consistency constraint for the forward physical field.

Specifically, the loss term of GI-ION can be expressed in the following form:
\begin{equation}
\mathcal{L}_{\text{data}}(\bm{m},\bm{d})
= \frac{1}{N} 
    \sum_{i=1}^{N} 
    \left\| \mathcal{F}_\theta (\boldsymbol{m}_i)(\boldsymbol{x}_\text{o}) - \boldsymbol{d}_i \right\|^2, 
    \label{eq:data-loss}
\end{equation}
\begin{equation}
\mathcal{L}_{\text{physics}}(\bm{m})
= \frac{1}{N} \sum_{i=1}^{N} \left( 
    \alpha \mathcal{L}_{\mathcal{N}} (\boldsymbol{m}_i) 
    + \beta \mathcal{L}_{\mathcal{B}} (\boldsymbol{m}_i) \right),
    \label{eq:physics-loss}
\end{equation}
where \(\alpha\) and \(\beta\) are weighting coefficients. The physics-informed loss comprises two terms, $\mathcal{L}_{\mathcal{N}}$ and $\mathcal{L}_{\mathcal{B}}$:

\begin{equation}
\mathcal{L}_{\mathcal{N}} (\boldsymbol{m}) =  \left\| \left[ \mathcal{N} \left( \mathcal{F}_\theta (\boldsymbol{m}); \boldsymbol{m} \right) \right] \left( \boldsymbol{x}_\text{total} \right) \right\|^2, 
\end{equation}

\begin{equation}
\mathcal{L}_{\mathcal{B}} (\boldsymbol{m}) =  \left\| \left[ \mathcal{B} \left( \mathcal{F}_\theta (\boldsymbol{m}); \boldsymbol{m} \right) \right] \left( \boldsymbol{x}_\text{total} \right) \right\|^2, 
\end{equation}
where $\mathcal{L}_{\mathcal{N}}$ and $\mathcal{L}_{\mathcal{B}}$ denote the physics-informed losses associated with the physical processes $\mathcal{N}$ and $\mathcal{B}$, respectively.

While this physics-based term mathematically acts as a regularizer in the joint objective, it imposes an observation-free prior on the predicted physical state field $\bm{u}$ through the forward governing equations, rather than directly penalizing the model parameters $\bm{m}$ as in conventional regularization (e.g., Tikhonov). Specifically, by minimizing the PDE residual, it enforces global physical consistency of $\bm{u}$, which in turn implicitly constrains $\bm{m}$ toward models that are compatible with the governing physical process.

\subsection{Novel independence loss}

Unlike standard INN training strategies that rely on kernel-based MMD to constrain the latent distribution (see Appendix~\ref{app:MMD}) \cite{ardizzoneanalyzing}, we introduce a novel independence loss because MMD can be unstable and computationally expensive in high dimensions. This loss explicitly minimizes statistical dependence between the latent variables and branch coefficients, providing a more robust constraint for data-limited geophysical scenarios. The output of the branch network $\mathcal{B}_\theta(\boldsymbol{m})$ includes latent noise $\boldsymbol{z}$ to characterize uncertainty in the inverse problem. Additional constraints are required for the forward output $p_\theta(\boldsymbol{b}, \boldsymbol{z})$ to satisfy the independence condition $p_\theta(\boldsymbol{b})p(\boldsymbol{z})$ in \cref{eq:independence}.

\begin{equation}\label{eq:independence}
    \left\{\begin{array}{l}
	p_\theta\left( \boldsymbol{b}, \boldsymbol{z}\right)=p_\theta\left(\boldsymbol{b}\right) p(\boldsymbol{z}), \\
	\boldsymbol{z} \sim \mathcal{N}(\mathbf{0},\mathbf{I}).
\end{array}\right. 
\end{equation}

Before defining the loss, we note that computing $p_\theta\left(\boldsymbol{b}\right)=\int p_\theta\left(\boldsymbol{b}, \boldsymbol{z}\right) \mathrm{d} \boldsymbol{z}$ requires multiple integrals, which makes independence difficult to measure directly using \cref{eq:independence}. However, the joint distribution $p_\theta\left(\boldsymbol{b}, \boldsymbol{z}\right)$ can be computed analytically from the model prior $p_0(\boldsymbol{m})$ and $\mathcal{B}_\theta$ through the inverse process

\begin{equation}
	p_\theta\left(\boldsymbol{b}, \boldsymbol{z}\right) = p_0\left(\mathcal{B}_\theta^{-1}(\boldsymbol{b}, \boldsymbol{z})\right) \left| \det 
    J_{\mathcal{B}_\theta^{-1}}(\boldsymbol{b}, \boldsymbol{z}) \right|. 
\end{equation}

Notice that a necessary condition for \cref{eq:independence} to hold is that for any $\left(\boldsymbol{b}, \boldsymbol{z}\right)$ and for any chosen new noise $\boldsymbol{z}^{*}$, we obtain

\begin{equation}\label{eq:ind-condition}
    \frac{p_\theta\left(\boldsymbol{b}, \boldsymbol{z}\right)}{p_\theta\left(\boldsymbol{b}, \boldsymbol{z}^{*}\right)}=\frac{p(\boldsymbol{z})}{p\left(\boldsymbol{z}^{*}\right)}, \quad \forall \boldsymbol{b}, \boldsymbol{z}, \boldsymbol{z}^{*} . 
\end{equation}

On points where the density ratios are defined, \cref{eq:ind-condition} is also sufficient for the independence relation in \cref{eq:independence}; Appendix~\ref{app:ind-loss} provides the corresponding argument. The first independence-loss term penalizes deviations from this ratio condition. For each training pair $(\boldsymbol{m}_i,\boldsymbol{d}_i)$, we define
\[
(\boldsymbol{b}_i,\boldsymbol{z}_i)=\mathcal{B}_{\theta}(\boldsymbol{m}_i),
\qquad
\boldsymbol{b}_i^{*}=\mathbf{T}_{\mathrm{o}}^{\dagger}\boldsymbol{d}_i.
\]
Thus, $\boldsymbol{b}_i$ and $\boldsymbol{b}_i^{*}$ are paired sample by sample rather than drawn independently from two marginal distributions. The empirical distribution of the data-projected coefficients is
\[
\widehat{p}_{\mathrm{proj}}(\boldsymbol{b})
:=
\frac{1}{N}\sum_{i=1}^{N}\delta_{\boldsymbol{b}_i^{*}}(\boldsymbol{b}).
\]
This empirical measure is not assumed to be an analytically available population density. The independence loss is

\begin{equation}\label{eq:ind-loss}
    \begin{aligned}
    \mathcal{L}_{\text{ind}}
            &= \mathcal{L}_{\text{ind}_1} + \mathcal{L}_{\text{ind}_2},\\
    \mathcal{L}_{\text{ind}_1}
            &= \mathbb{E}_{\substack{(\boldsymbol{b},\boldsymbol{z})\sim p_\theta(\boldsymbol{b},\boldsymbol{z})\\
                    \boldsymbol{z}^{*}\sim p(\boldsymbol{z})}}
               \left[\left(
               \log\frac{p_\theta(\boldsymbol{b},\boldsymbol{z})}{p_\theta(\boldsymbol{b},\boldsymbol{z}^{*})}
               -\log\frac{p(\boldsymbol{z})}{p(\boldsymbol{z}^{*})}
               \right)^2\right],\\
    \mathcal{L}_{\text{ind}_2}
            &= \frac{1}{N}\sum_{i=1}^{N}
               \left[\log\frac{p_\theta(\boldsymbol{b}_i,\boldsymbol{z}_i)}
               {p_\theta(\boldsymbol{b}_i^{*},\boldsymbol{z}_i)}\right]^2.
    \end{aligned}
\end{equation}
For a general observation vector, the data-projected coefficient is calculated as

\begin{equation}\label{eq:LQ}
\boldsymbol{b}^{*}(\boldsymbol{d}) =\left[\left(\mathcal{T}_{\theta}\left(x_\text{o}^{1}\right), \ldots, \mathcal{T}_{\theta}\left(x_\text{o}^{M_\text{o}}\right)\right)^{\top}\right]^\dagger\left(d^1, \ldots, d^{M_\text{o}}\right)^{\top},
\end{equation}
where $\dagger$ denotes the Moore--Penrose inverse.
In finite-sample training, the ratio constraint represented by $\mathcal{L}_{\text{ind}_1}$ may not be fully satisfied because of limited network capacity and optimization error. We therefore use $\mathcal{L}_{\text{ind}_2}$ as an empirical stabilization term. It penalizes residual log-density differences between the paired learned and data-projected coefficients at the same latent variable $\boldsymbol{z}_i$. It is not interpreted as an exact divergence, a marginal-distribution matching objective, or a guarantee that $p_\theta(\boldsymbol{b})=\widehat{p}_{\mathrm{proj}}(\boldsymbol{b})$.
We demonstrate the effect of the independence loss in the experiment described in Section~\ref{sec:surface-wave}.

In summary, we can train GI-ION by minimizing the following combined loss. 
\begin{equation}\label{eq:total-loss}
 \mathcal{L}=\lambda_1\mathcal{L}_{\text{physics}} + \lambda_2\mathcal{L}_{\text{data}} + \lambda_3\mathcal{L}_{\text{ind}}.
\end{equation}

By minimizing the combined loss, GI-ION effectively integrates data-driven and physics-based constraints while ensuring the invertibility necessary for robust geophysical inversion.

\begin{algorithm}[ht]
\caption{GI-ION training procedure}
\label{alg:training}
\begin{algorithmic}[1]
\STATE \textbf{Input:} Training dataset $\{\mathbf{m}_i, \mathbf{d}_i\}_{i=1}^N$, query points $\{\boldsymbol{x}_\text{total}\}$, and hyperparameters $\lambda_1, \lambda_2, \lambda_3$

\STATE Initialize network parameters $\theta$
\FOR{epoch $= 1$ to $E$}
    \FOR{batch $(\boldsymbol{m}_b, \boldsymbol{d}_b)$ in the dataset}
        \STATE Compute $\boldsymbol{t}(\boldsymbol{x}) = \mathcal{T}_\theta(\boldsymbol{x})$
        \STATE Compute $[\boldsymbol{b},\boldsymbol{z}] = \mathcal{B}_\theta(\boldsymbol{m}_b)^{\top}$
        \STATE Compute $\boldsymbol{u} = \mathcal{F}_{\theta}(\boldsymbol{m}_b)(\boldsymbol{x})$ (\cref{eq:deeponet-expansion})
        \STATE Compute data loss $\mathcal{L}_{\text{data}}$ (\cref{eq:data-loss})

        \STATE Compute physics loss $\mathcal{L}_{\text{physics}}$ (\cref{eq:physics-loss})

        \STATE Sample $\boldsymbol{z}^{*}\sim\mathcal{N}(\mathbf{0},\mathbf{I})$ and compute $\boldsymbol{b}^*$ (\cref{eq:LQ})
        \STATE Compute the joint densities $p_\theta\left(\boldsymbol{b}, \boldsymbol{z}\right)$, $p_\theta\left(\boldsymbol{b}^*, \boldsymbol{z}\right)$, $p_\theta\left(\boldsymbol{b}, \boldsymbol{z}^*\right)$ (\cref{eq:ind-loss})
        
        \STATE Compute independence loss $\mathcal{L}_{\text{ind}}$ (\cref{eq:ind-loss})

        \STATE Total loss $\mathcal{L} = \lambda_1\mathcal{L}_{\text{physics}} + \lambda_2\mathcal{L}_{\text{data}} + \lambda_3\mathcal{L}_{\text{ind}}$ (\cref{eq:total-loss})
        \STATE Update $\theta$  
    \ENDFOR

\ENDFOR
\STATE \textbf{Output:} Trained network $\mathcal{F}_\theta$ (branch $\mathcal{B}_\theta$ and trunk $\mathcal{T}_\theta$)
\end{algorithmic}
\end{algorithm}

\subsection{Solving inverse problems with GI-ION}

The trained GI-ION enables fast conditional sampling. With an appropriately selected trunk-output dimension $P$, we assume that the retained trunk functions $\{t_j\}_{j=1}^{P}$ form a linearly independent representation of the learned solution space in $L^2(\mathcal{D})$. For a given observation set, the coefficients are estimated using the discrete matrix $\mathbf{T}_{\mathrm{o}}$. When $\mathbf{T}_{\mathrm{o}}$ has full column rank, $\boldsymbol{b}^{*}(\boldsymbol{d})=\mathbf{T}_{\mathrm{o}}^{\dagger}\boldsymbol{d}$ is the unique least-squares estimate; otherwise, the Moore--Penrose inverse returns the minimum-norm least-squares estimate. Using the inverse branch map, conditional samples are obtained from
\begin{equation}
\boldsymbol{m}=\mathcal{B}_\theta^{-1}\left(\boldsymbol{b}, \boldsymbol{z}\right), 
\end{equation}
where $\boldsymbol{b}=\boldsymbol{b}^{*}(\boldsymbol{d})$ and $\boldsymbol{z}\sim p(\boldsymbol{z})$. This sampling rule produces samples from the learned conditional distribution $q_\theta(\cdot\mid\boldsymbol{d})$ in \cref{eq:conditional-pushforward}, which approximates the Bayesian posterior. The complete procedure is outlined in \cref{alg:gi-ion-inference}.

\begin{algorithm}[ht]
\caption{Solving an inverse problem with GI-ION}
\label{alg:gi-ion-inference}
\begin{algorithmic}[1]
\STATE \textbf{Input:} Trained GI-ION $\left(\mathcal{B}_\theta, \mathcal{T}_\theta\right)$, observation vectors $\left\{\boldsymbol{d}_{i}\right\}_{i=1}^N$, and number of inversion samples $M$
\FOR{$i = 1$ \textbf{to} $N$}
\STATE Calculate $\boldsymbol{b}_i$ by \cref{eq:LQ}  
\FOR{$j = 1$ \textbf{to} $M$}
\STATE Draw $\boldsymbol{z}^j \sim p(\boldsymbol{z})$
\STATE Calculate $\boldsymbol{m}^{j}_i = \mathcal{B}_\theta^{-1}(\boldsymbol{b}_i, \boldsymbol{z}^j)$
\ENDFOR
\ENDFOR
\STATE \textbf{Output:} Approximate posterior samples $\left\{\boldsymbol{m}^{j}_i\right\}_{j=1}^{M}$ for each observation vector $\boldsymbol{d}_{i}$
\end{algorithmic}
\end{algorithm}

\subsection{Related work and novelty}

To position GI-ION within inversion based on deep learning, we distinguish our contributions from existing approaches in terms of the inference paradigm, operator architecture, and training strategy. Standard CNN-based methods, such as U-Net, are typically restricted to estimation on fixed discretization grids and inherently fail to assess solution reliability \cite{puzyrev2019deep, oh2018salt, puzyrev2021inversion, poddar2025inversion}. Conversely, although PINNs allow flexible discretization while promoting physical consistency, they typically operate as single-instance solvers, necessitating computationally expensive retraining for each new observation \cite{rasht2022physics, waheed2021pinntomo, baldan2023physics}. GI-ION overcomes these limitations through a physics-informed operator framework. Once trained, it enables real-time probabilistic inference for new data without retraining.
Unlike the conventional DeepONet architecture \cite{kaltenbach2023semi, lu2021learning,wang2021learning}, the proposed invertible branch network provides an invertible mapping between model parameters and the coefficient--latent representation. This allows the network to learn the forward physical operator while supporting probabilistic reverse inference. The independence-oriented loss penalizes residual dependence between the branch coefficients and latent variables. In the experiments considered here, it yielded lower residual dependence than the MMD-regularized baseline \cite{ardizzoneanalyzing} and supported approximate conditional sampling.

\section{Numerical experiments}
\label{sec:numerical-experiments}

In this section, we validate GI-ION using a series of numerical experiments. We first demonstrate the effectiveness of the independence loss in the surface-wave dispersion inversion problem and benchmark its performance against the standard INN with MMD loss \cite{ardizzoneanalyzing, zhang2021bayesian}.
Next, we apply GI-ION to inverse problems arising from 2D synthetic random-field travel-time tomography to evaluate the physics-informed loss mechanism within the GI-ION architecture.
Finally, to assess the generalizability of GI-ION to realistic models, we apply travel-time tomography to a seismic-slice model.

\subsection{1D surface-wave dispersion inversion problem}
\label{sec:surface-wave}

\begin{figure}[!t]
	\centering
	\includegraphics[width=0.9\textwidth]{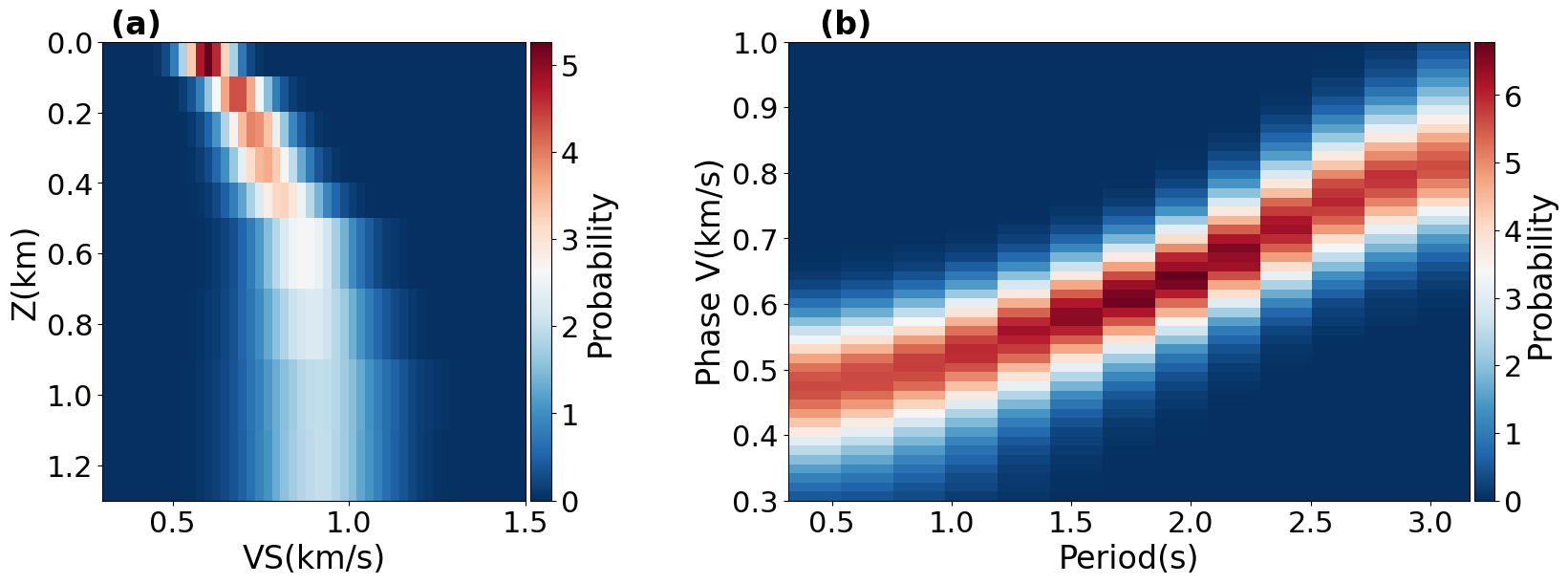}
	\caption{Prior distributions of (a) shear-velocity models and (b) surface-wave dispersion data.}
	\label{fig:surface-wave-distributions}
\end{figure}

To evaluate the effectiveness of the proposed independence loss, we first consider a 1D surface-wave dispersion inversion problem. The subsurface is discretized into ten homogeneous layers, with layer thicknesses of $0.1~\mathrm{km}$ in the shallow region ($<0.5~\mathrm{km}$) and $0.2~\mathrm{km}$ in the deeper region ($>0.5~\mathrm{km}$). The shear-velocity model $\boldsymbol{m}$ to be estimated is generated from a normal prior distribution.
The P-wave velocity $V_p$ and density are calculated from the shear velocity using $V_p = 1.16\boldsymbol{m} + 1.36$ \cite{castagna1985relationships} and $\rho = 1.74V_p^{0.25}$ \cite{brocher2005empirical}.
Surface-wave dispersion curves $\boldsymbol{d}$ are computed at a period sampling interval of $0.1~\mathrm{s}$ over the range $0.7$--$2.0~\mathrm{s}$ (\cref{fig:surface-wave-distributions}).

\begin{figure}[!t]
	\centering
	\includegraphics[width=0.9\textwidth]{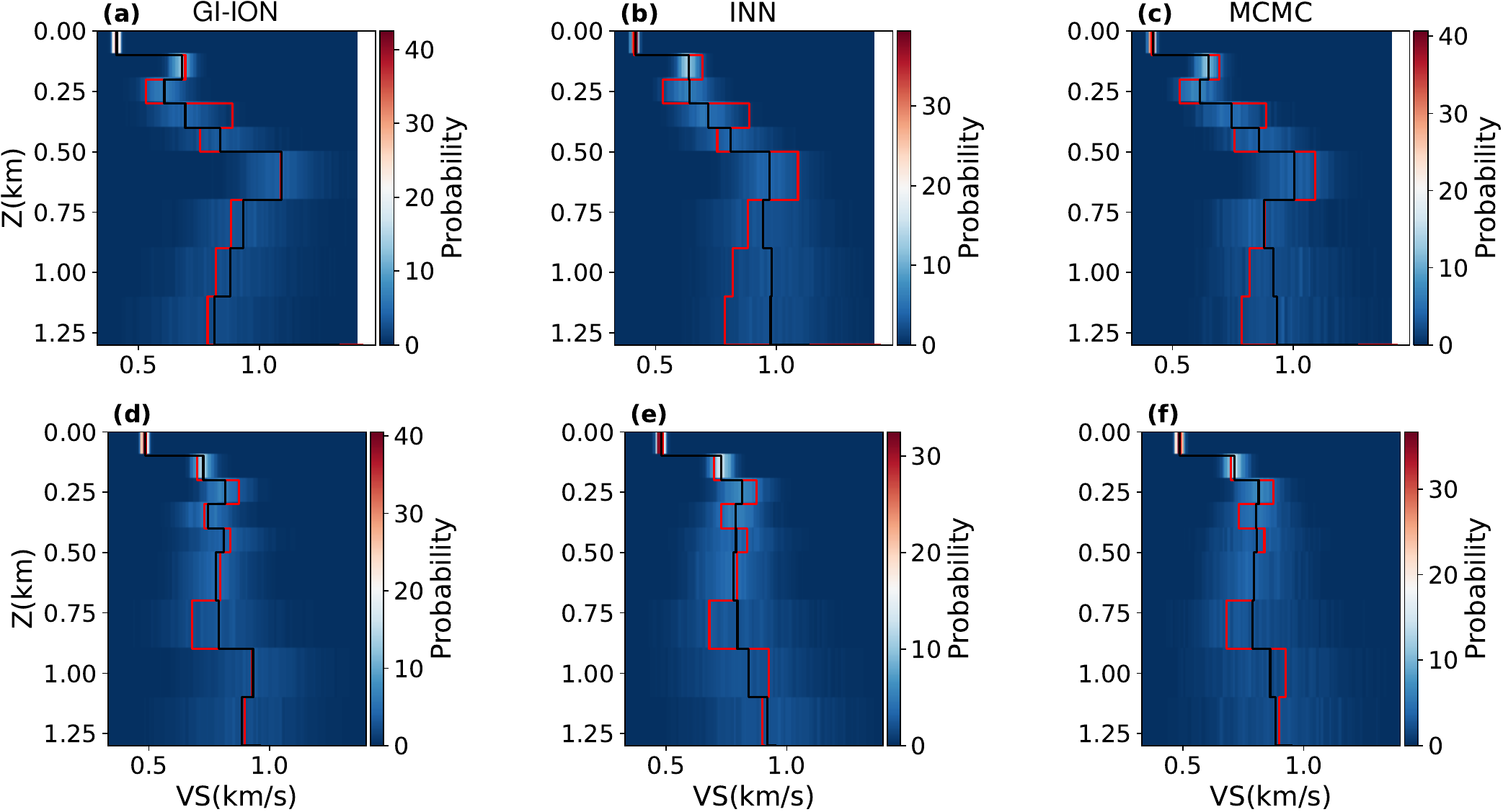}
\caption{Posterior estimates for two shear-velocity models obtained using different methods. Each row corresponds to one model, and each column corresponds to one method: (a, d) GI-ION, (b, e) standard INN, and (c, f) MCMC. Red lines denote true shear velocities, and black lines denote posterior mean velocities.}
	\label{fig:surface-wave-posterior-samples}
\end{figure}

In this low-dimensional setting, the neural basis functions can be simplified into orthogonal standard bases, effectively reducing GI-ION to a standard INN. In this case, the branch coefficients are equivalent to the observation data, i.e., $\boldsymbol{d} = \boldsymbol{b}$. To compare the effects of the MMD and independence losses, the standard INN employs a composite loss function consisting of a forward loss, a bidirectional MMD loss, and a maximum-likelihood loss \cite{ardizzoneanalyzing, zhang2021bayesian}. In contrast, the GI-ION trunk network is simplified to a standard orthogonal basis and uses a similar loss function, with the MMD loss replaced by the independence loss.
The MMD kernel uses the inverse multiquadratic function (see Appendix~\ref{app:MMD}), which has heavier tails than the Gaussian kernel, leading to improved robustness against outliers \cite{tolstikhin2018wasserstein}. Specific experimental details and hyperparameter settings are provided in Appendix~\ref{app: implementation details}. The posterior estimates obtained from different methods are presented in \cref{fig:surface-wave-posterior-samples}.

\begin{table}[!t]
    \centering
    \begin{threeparttable}
    \caption{Quantitative comparison at different noise levels.}
    \label{tab:surface-wave-noise}
        \setlength{\tabcolsep}{4pt}
        \begin{tabularx}{\textwidth}{l *{6}{>{\centering\arraybackslash}X}}
            \toprule
		Metrics & \multicolumn{3}{c}{MSE} & \multicolumn{3}{c}{CRPS}  \\
		\cmidrule(lr){2-4} \cmidrule(lr){5-7}  
		Noise level & 1\% & 3\% & 10\%  & 1\% & 3\% & 10\%  \\
		\midrule
		GI-ION &  \textbf{0.2387} &  \textbf{0.2431} &  \textbf{0.2802} & \textbf{0.05454}   & \textbf{0.05761}  & \textbf{0.06254}   \\
		INN  & 0.2711  & 0.2806 &  0.3074 & 0.05724  &  0.05963  & 0.06441  \\
        MCMC & 0.2831  & 0.2923 &  0.3187 & 0.05802  &  0.06047  & 0.06539   \\
            \bottomrule
        \end{tabularx}
    \begin{tablenotes}
            \footnotesize
        \item \textbf{Note:} GI-ION and standard INN were trained with $20\,000$ training samples. Metrics were averaged over $10\,000$ test samples. Best values are shown in bold.
    \end{tablenotes}
    \end{threeparttable}
\end{table}

\begin{figure}[!t]
	\centering
	\includegraphics[width=0.9\textwidth]{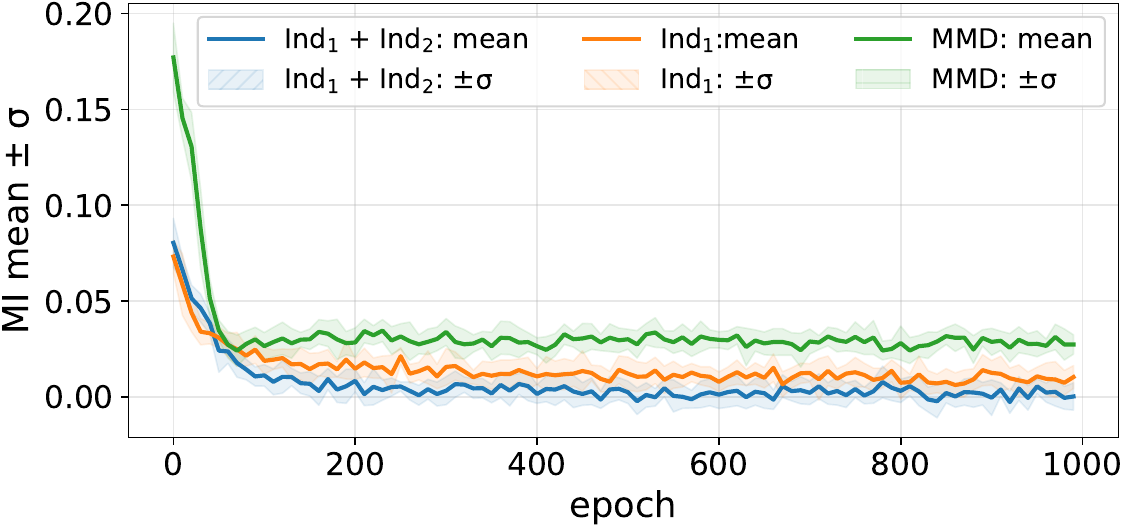}
    \caption{Mutual-information trajectories for different regularization terms in a surface-wave dispersion experiment.}
	\label{fig:mi-decrease}
\end{figure}

As shown in \cref{eq:independence}, the joint distribution requires both Gaussianity of $\boldsymbol{z}$ and independence between $\boldsymbol{b}$ and $\boldsymbol{z}$.
To verify Gaussianity, we extracted $\boldsymbol{z}$ from $10\,000$ posterior samples of the test set and computed the covariance matrix and one-dimensional marginals.
The covariance-matrix diagnostics show that the covariance matrix of latent variables obtained with independence-loss regularization is approximately an identity matrix, whereas the MMD loss introduces interdimensional correlations, which in turn generate spurious associations.
The histograms of randomly selected latent dimensions further indicate that the latent-noise output regularized by the independence loss is approximately Gaussian, whereas the MMD loss results in a more peaked (leptokurtic) distribution.
The more peaked latent distribution produced by the MMD-regularized model leads to overestimated latent variability when a Gaussian reference distribution is used for inversion, thereby widening the resulting uncertainty estimates.
Furthermore, the Wasserstein distance metric (defined in Appendix~\ref{app:Metrics}) provides additional confirmation that the latent noise under independence-loss regularization exhibits improved Gaussianity.
To verify independence, we calculated the mutual information metric (defined in Appendix~\ref{app:Metrics}) between $\boldsymbol{b}$ and $\boldsymbol{z}$ from $10\,000$ posterior samples of the test set. The results show that the independence loss enforces lower mutual information than the MMD loss and remains more robust under small-sample conditions. In addition, the mutual information trajectory during training (\cref{fig:mi-decrease}) shows that the independence loss provides effective regularization and accelerates convergence. Mutual information between the branch coefficients $\boldsymbol{b}$ and latent variables $\boldsymbol{z}$ was recorded every ten epochs over $1\,000$ training iterations using $20\,000$ training samples. Each experiment was repeated ten times, and the mean and standard deviation of the mutual information were computed. To assess the effect of noise level on performance, we evaluated all competing methods under three noise conditions. The quantitative results in \cref{tab:surface-wave-noise} show that the posterior estimation error increases with noise level for all methods and that the proposed method consistently maintains superior accuracy.

\subsection{2D random-field travel-time tomography}
\label{sec:random-field-tomography}

\begin{figure}[!t]
	\centering
  	\includegraphics[width=0.9\textwidth]{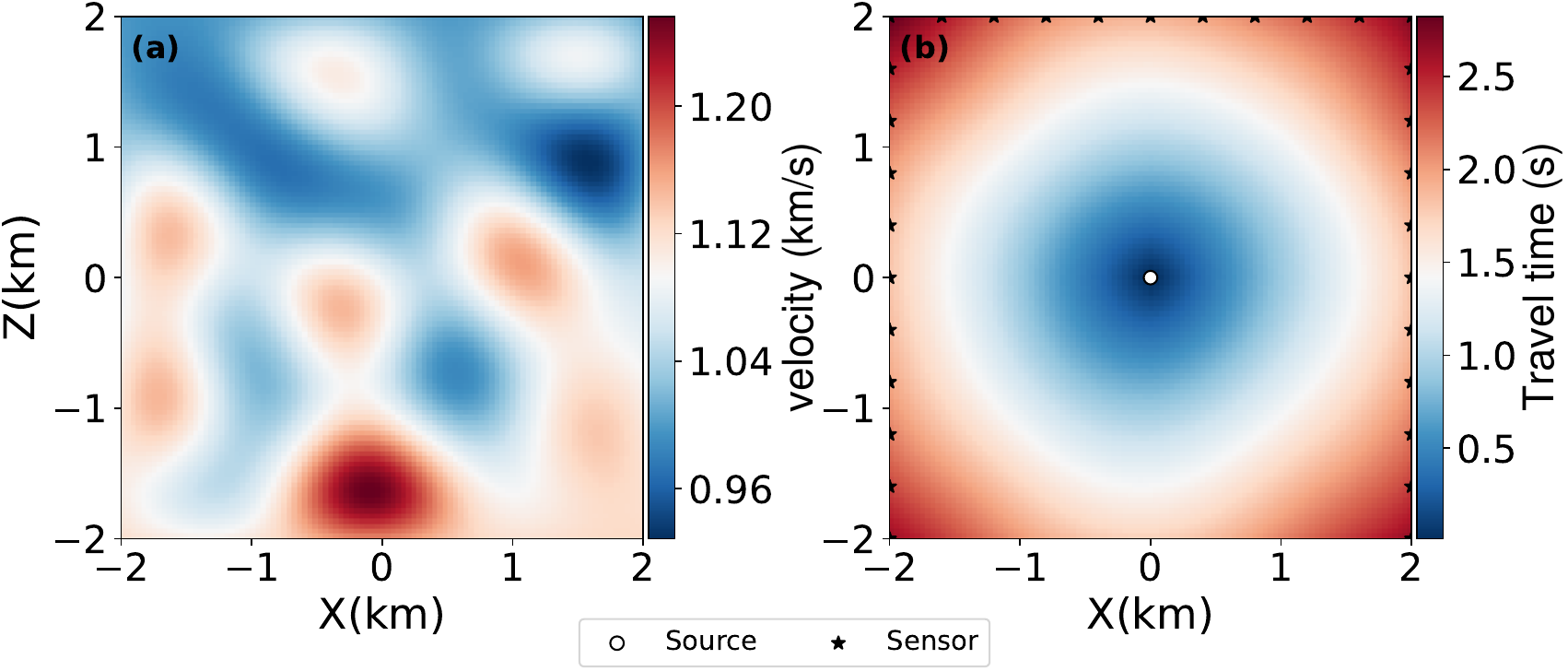}
	\caption{(a) Example velocity model represented by a Gaussian random field and (b) corresponding travel-time field with source and sensor locations.}
	\label{fig:random-field-travel-time}
\end{figure}

In geophysical inverse problems, data-driven models struggle to capture the inherent uncertainty and yield unreliable results when physical constraints are absent and data are limited. To systematically validate the efficacy of the physics-informed loss mechanism within the GI-ION architecture, we conducted a series of 2D random-field travel-time tomography inversion experiments. As depicted in \cref{fig:random-field-travel-time}(a), the computational domain comprises a $4\ \mathrm{km} \times 4\ \mathrm{km}$ square discretized into a structured $101 \times 101$ grid with spatial resolution $\Delta x = 0.04\ \mathrm{km}$.
The velocity-field model $\boldsymbol{m} = V_r + V_g$ to be inverted comprises a Gaussian random field $V_r$, constructed using a truncated Karhunen--Loève expansion retaining 20 dominant modes, and a random background-gradient field $V_g$ with $ \nabla V_g \sim \mathcal{U}[0.03, \ 0.06]\ \mathrm{km/s}$ to characterize epistemic uncertainty.
The initial seismic source is positioned at the center of the rectangular domain (\cref{fig:random-field-travel-time}(b)), while sensors are placed at $0.2~\mathrm{km}$ intervals along the boundary, yielding a total of 80 sensors.
Synthetic travel times for the random-field models were computed using the fast marching method based on the eikonal equation \citep{sethian1996fast}. To introduce aleatory uncertainty and simulate observational noise, Gaussian noise with a mean of $0$ and a standard deviation of \(5 \times 10^{-3}~\mathrm{s}\) was added to the synthetic travel-time data \cite{zhang2021introduction, linde2015geological, gosselin2022review}. Specific experimental details and hyperparameter settings are provided in Appendix~\ref{app: implementation details}.

\begin{figure}[!t]
	\centering
	\includegraphics[width=0.9\textwidth]{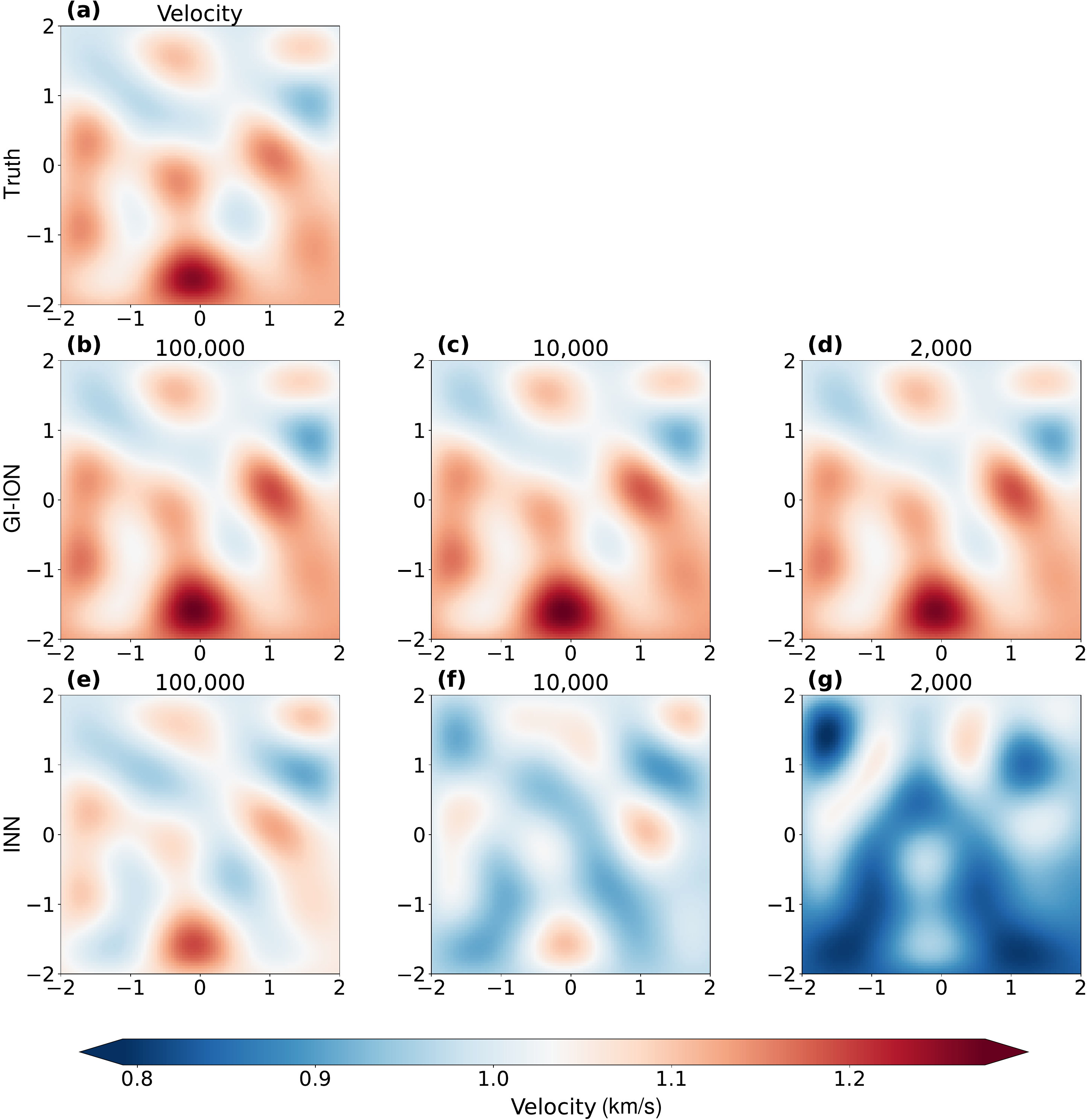}
	\caption{Posterior means obtained with different training-set sizes: (a) ground truth; (b--d) GI-ION; and (e--g) standard INN.}
	\label{fig:random-field-posterior-means}
\end{figure}

\begin{figure}[ht]
	\centering
	\includegraphics[width=0.82\textwidth]{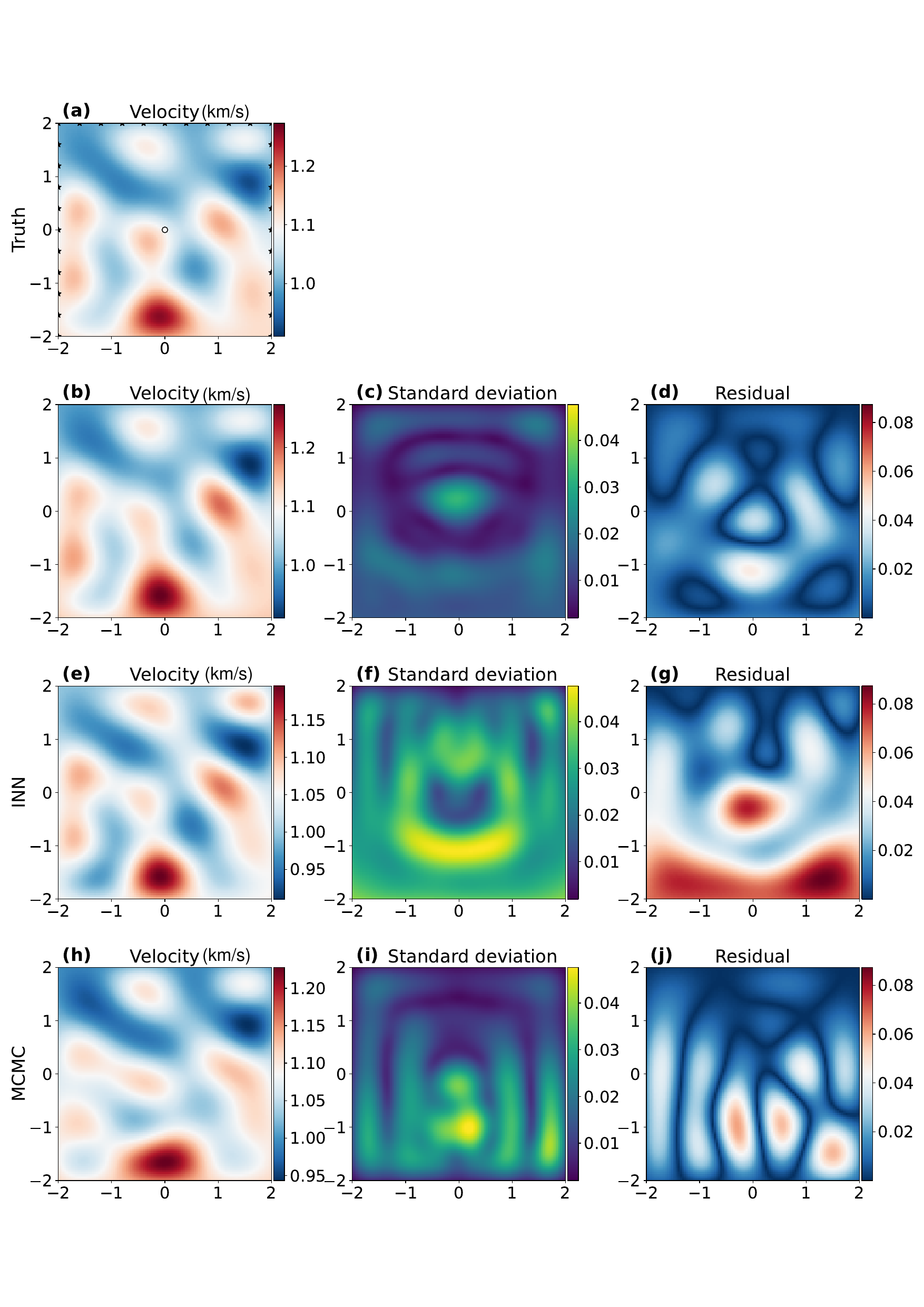}
	\caption{Posterior estimates from three methods for one test sample:
(a) ground truth; (b--d) GI-ION; (e--g) standard INN; and (h--j) MCMC.
Columns within each method row show posterior mean, posterior standard
deviation, and residual between posterior mean and ground truth,
respectively. GI-ION and standard INN were trained with $100\,000$ samples.}
        \label{fig:random-field-posterior-estimates}
\end{figure}

\Cref{fig:random-field-posterior-means} shows posterior means under different training-set sizes, and \cref{fig:random-field-posterior-estimates} further compares the posterior mean, standard deviation, and residual. GI-ION consistently provides more accurate reconstructions than the standard INN, with the advantage becoming more pronounced in data-scarce settings. The standard INN performs reasonably near sensor locations but deteriorates in unobserved regions, whereas GI-ION benefits from physics-informed regularization that stabilizes the inversion beyond the observed domain. MCMC shows comparatively larger uncertainty and reconstruction error in this setting.

To establish quantitative performance benchmarks, we employed two evaluation metrics (\cref{tab:random-field-metrics}): the structural similarity index measure (SSIM) and the signal-to-noise ratio (SNR), both defined in Appendix~\ref{app:Metrics}. Originally developed for image reconstruction, these metrics are introduced here to assess reconstruction fidelity in high-dimensional geophysical inverse problems. Each metric was computed using $5\,000$ retained samples from the corresponding conditional distribution. For MCMC, samples were retained after burn-in and thinning. The results show that GI-ION consistently outperforms other methods across all dataset sizes, with particularly strong performance retention under data-limited conditions. This improvement demonstrates that the introduced physics-based loss effectively regularizes the inversion by incorporating physical information.

\begin{table}[!t]
    \centering
    \begin{threeparttable}
      \caption{Quantitative comparison among three inversion methods.}
        \label{tab:random-field-metrics}
        \begin{tabularx}{\textwidth}{l *{6}{>{\centering\arraybackslash}X}}
            \toprule
		Metrics & \multicolumn{3}{c}{SSIM} & \multicolumn{3}{c}{SNR}  \\ 
		\cmidrule(lr){2-4} \cmidrule(lr){5-7}  
		$N_{\mathrm{train}}$ & $100\,000$ & $10\,000$ & $2\,000$  & $100\,000$ & $10\,000$ & $2\,000$   \\
		\midrule
		GI-ION &  \textbf{0.9905} &  \textbf{0.9854} &  \textbf{0.9785} & \textbf{39.91}   & \textbf{36.81}  & \textbf{33.60}   \\ 
		INN & 0.9448 & 0.9411 & 0.8458 & 28.44 & 27.73 &  24.21  \\
            MCMC &  0.9571 & N/A & N/A & 34.03 & N/A & N/A \\
            \bottomrule
        \end{tabularx}
    \begin{tablenotes}
            \footnotesize
            \item \textbf{Note:} GI-ION and standard INN were trained with different training-set sizes. MCMC was run for $220\,000$ iterations, with the first $120\,000$ iterations discarded as burn-in and a thinning interval of 20 iterations applied thereafter. Metrics were averaged over the $10\,000$-sample test set. Best values are shown in bold.
    \end{tablenotes}
    \end{threeparttable}
\end{table}

\subsection{Seismic-slice travel-time tomography}
\label{sec:seismic-slice}

\begin{figure}[ht]
	\centering
	\includegraphics[width=0.9\textwidth]{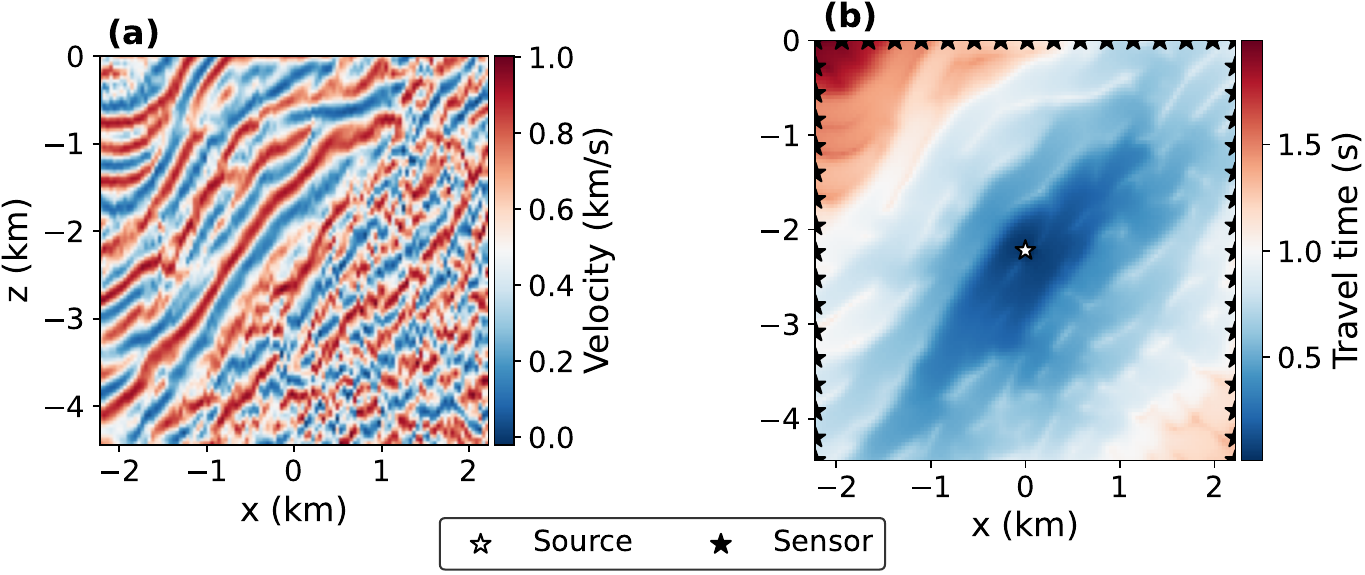}
	\caption{(a) Example 2D cross-sectional velocity-field model and (b) corresponding travel-time field with source and sensor locations.}
	\label{fig:seismic-slice-travel-time}
\end{figure}

To assess the generalizability of GI-ION to realistic complex models, we use a 2D seismic slice extracted from a 3D seismic volume to construct the dataset \cite{sheng2025seismic}. The 2D seismic-slice model, shown in \cref{fig:seismic-slice-travel-time}(a), has a grid spacing of $40~\mathrm{m}$ and comprises 112 nodes in each dimension. As shown in \cref{fig:seismic-slice-travel-time}(b), wave propagation originates from a central point in the study area. Boundary observation locations are distributed along the upper and lateral boundaries of the domain, with a uniform spacing of $0.28~\mathrm{km}$ between adjacent stations.

Constructing a suitable prior distribution directly in a high-dimensional sample space is often difficult. Deep generative models provide a way to learn complex high-dimensional distributions and construct a deep generative prior \cite{pan2021exploiting, jalal2021robust, xia2023vi}. Accordingly, we use a pretrained variational autoencoder (VAE) to learn the complex distribution of the sample models and obtain a low-dimensional representation; the detailed VAE architecture is provided in Appendix~\ref{app: DGP network}. This deep generative prior systematically quantifies the uncertainty associated with reconstruction errors during model inference, thereby improving the reliability and robustness of the reconstructions. The resulting uncertainty estimates support a more comprehensive evaluation of the spatial distribution and magnitude of confidence across the model prior.
Specific experimental details and hyperparameter settings are provided in Appendix~\ref{app: implementation details}.

\begin{figure}[H]
	\centering
	\includegraphics[width=\textwidth]{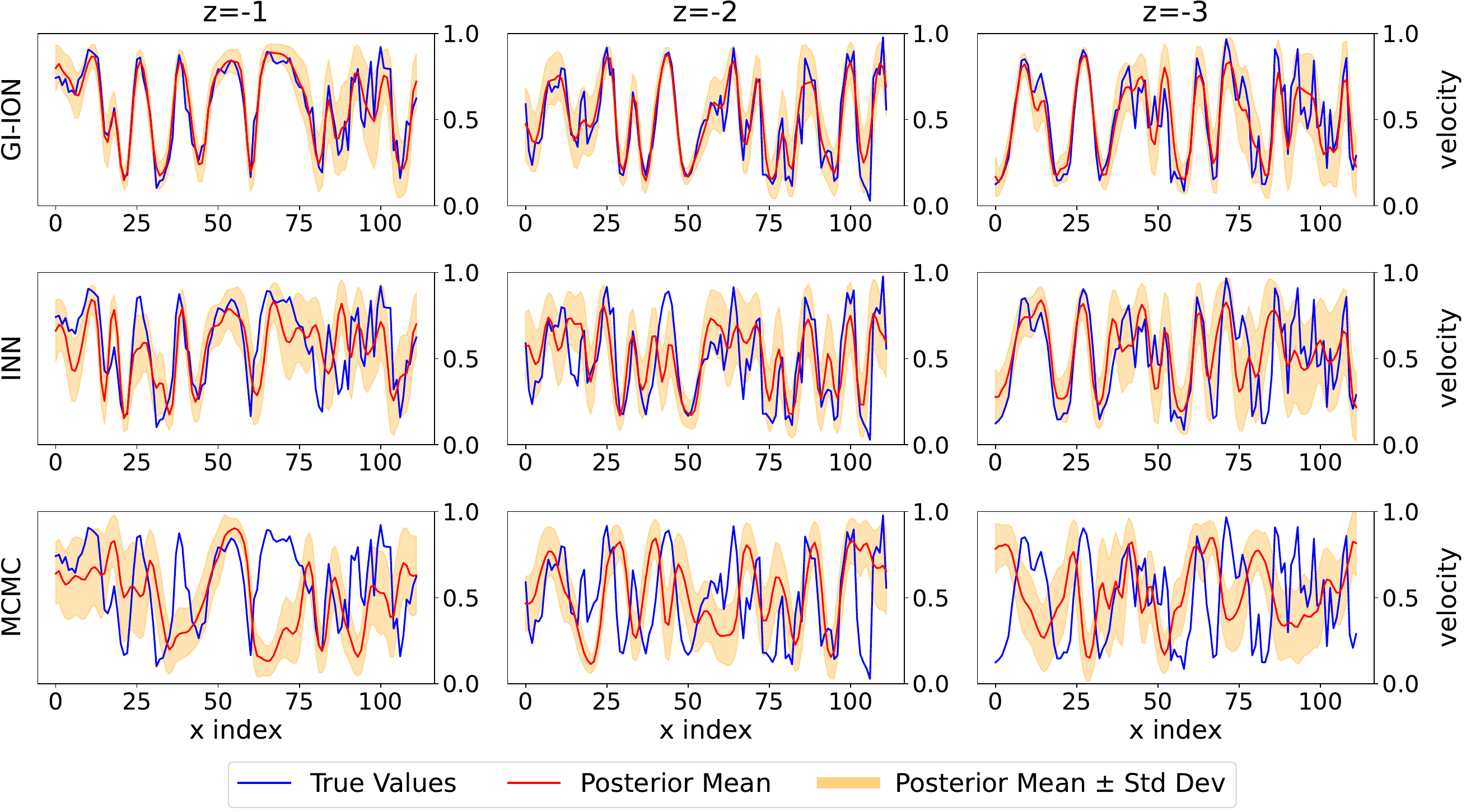}
	\caption{Posterior-estimate profiles obtained using different methods
at three depths. Blue lines denote true values, red lines denote
posterior means, and yellow bands indicate posterior mean plus or
minus one posterior standard deviation.}
	\label{fig:seismic-slice-profiles}
\end{figure}

\Cref{fig:seismic-slice-posterior} compares the inversion results on the seismic-slice example. All methods recover the dominant large-scale trend, but GI-ION better preserves the main structural pattern while providing more reliable uncertainty estimates. The standard INN reconstructs boundary-adjacent structures reasonably well but introduces artifacts in the interior, whereas MCMC shows substantially poorer agreement with the true model in this high-dimensional sparse-observation setting. The profile comparison in \cref{fig:seismic-slice-profiles} and the quantitative results in \cref{tab:seismic-slice-metrics} support the same conclusion.

\begin{figure}[ht]
	\centering
	\includegraphics[width=0.9\textwidth]{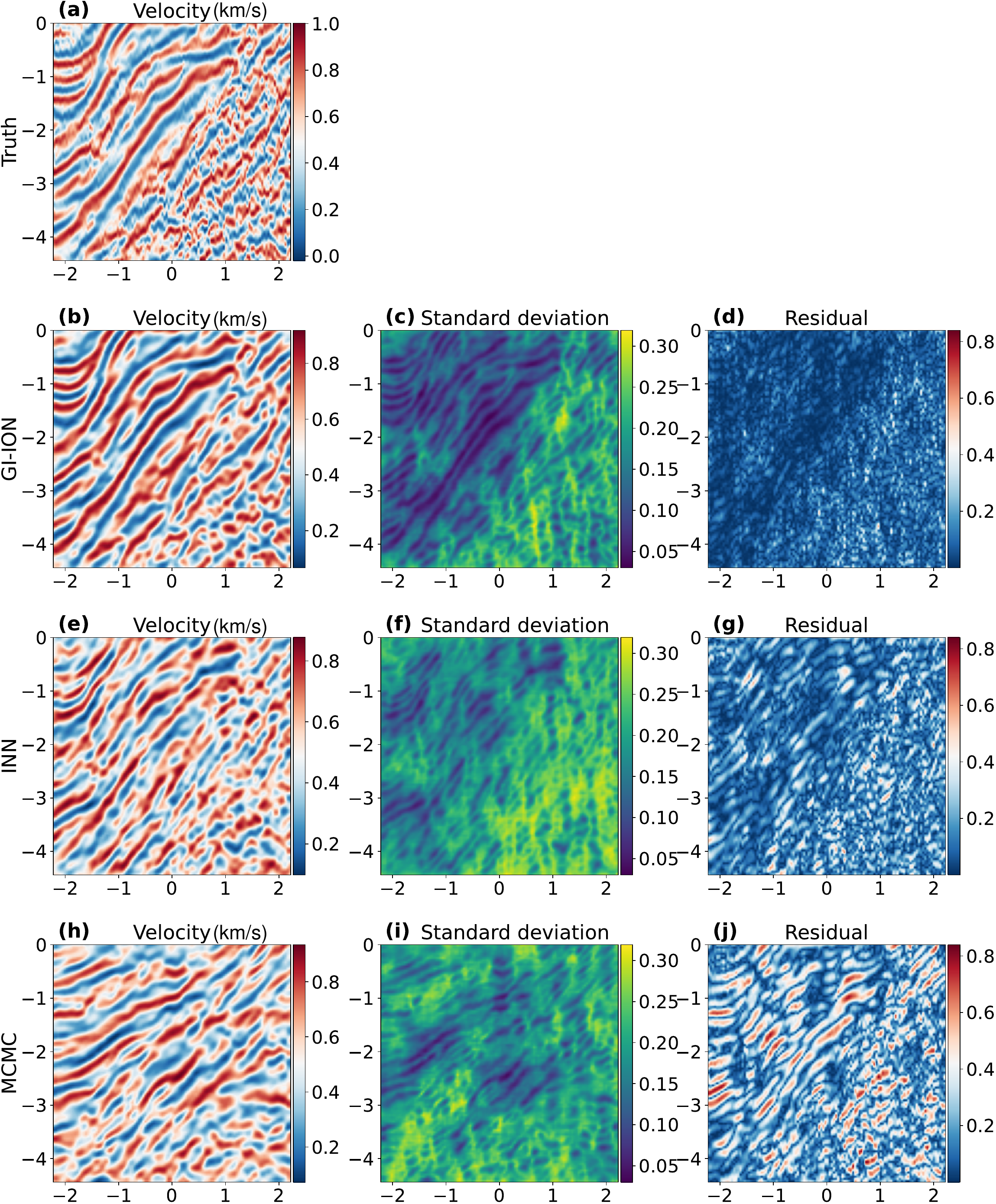}
    \caption{Posterior estimates from three methods for one test sample: (a) true velocity field; (b, e, h) posterior means; (c, f, i) posterior standard deviations; and (d, g, j) errors between posterior means and true velocity field.}
	\label{fig:seismic-slice-posterior}
\end{figure}

\begin{table}[!t]
\centering
\begin{threeparttable}
    \caption{Quantitative comparison among three inversion methods.}
    \label{tab:seismic-slice-metrics}
    \begin{tabular*}{\textwidth}{@{\extracolsep{\fill}}lcc@{}}
        \toprule
        \textbf{Method} & \textbf{SSIM} & \textbf{SNR} \\
        \midrule
        GI-ION & \textbf{0.7181} & \textbf{12.9388} \\
        INN    & 0.6095          & 11.0459          \\
        MCMC   & 0.0453          & 5.4632           \\
        \bottomrule
    \end{tabular*}
    \begin{tablenotes}
        \footnotesize
        \item \raggedright \textbf{Note:} Metrics were averaged over the test set. Best values are shown in bold.
    \end{tablenotes}
\end{threeparttable}
\end{table}

\clearpage

\section{Conclusions}
\label{sec:conclusions}

This study introduces GI-ION, a geophysics-informed invertible operator network that integrates invertible neural networks with geophysical knowledge for efficient Bayesian geophysical inversion. GI-ION introduces a latent variable to represent uncertainty in the inverse problem and uses an independence-oriented loss to regularize the coefficient--latent representation. Numerical experiments demonstrate accurate inversion and efficient sampling from the learned conditional approximation under the considered sparse- and limited-data conditions.

In future work, we intend to apply this architecture to 3D elastic-wave equations and real field data. This will require engineering adaptations to align real-world measurements with the theoretical assumptions of the GI-ION framework, thereby enabling robust inversion in environments with complex data.

\FloatBarrier

\subsection*{Data availability} The data supporting this study are available in Zenodo, Version v1, at \url{https://doi.org/10.5281/zenodo.19778388} \citep{zhang2026giiondataset}. Some data were obtained from previously published third-party sources cited in the manuscript. These third-party data remain subject to the original providers' licensing and distribution terms and are therefore not redistributed by the authors.

\subsection*{Declaration of generative AI and AI-assisted technologies
in the manuscript preparation process}

During the preparation of this work, the authors used ChatGPT and
OpenAI Codex to assist with language editing, manuscript organization,
and LaTeX-format checking. After using these tools, the authors reviewed
and edited the content as needed and take full responsibility for the
content of the manuscript.

\subsection*{Computer code availability}

The source code is publicly available at \url{https://github.com/ProfGuan/Geophysics-informed-invertible-operator-networks/tree/main}. The code is released under the MIT License.

\subsection*{CRediT authorship contribution statement}

Yangwen Zhang: Conceptualization, Methodology, Software, Validation, Visualization, Writing -- original draft, Writing -- review \& editing. Xintong Wang: Conceptualization, Methodology, Software, Validation, Writing -- original draft, Writing -- review \& editing. Xiaofei Guan: Conceptualization, Methodology, Writing -- original draft, Writing -- review \& editing, Funding acquisition. Peng Yu: Conceptualization, Writing -- review \& editing.

\subsection*{Declaration of competing interest}

The authors declare that they have no known competing financial interests or personal relationships that could have appeared to influence the work reported in this paper.

\subsection*{Funding}

This work was supported by the National Natural Science Foundation of China (Grant No. 12271409), the Foundation of National Key Laboratory of Computational Physics, the Fundamental Research Funds for the Central Universities, and the Basic Research Project of CNPC (Grant No. 2023ZZ05).

\FloatBarrier

\clearpage
\appendix
\numberwithin{equation}{section}
\counterwithin{table}{section}
\counterwithin{figure}{section}
\counterwithin{theorem}{section}
\input{appendix}
\FloatBarrier
\clearpage
\bibliographystyle{plainnat}
\bibliography{references}
\end{document}

%% file: appendix.tex
\section{Asymptotic analysis}
\label{app:ind-loss}
This section clarifies the density-ratio condition used by the first independence-loss term and the empirical role of the second term. It also states the pushforward conditional distribution used in Eqs.~\eqref{eq:posterior_sampling_rule}--\eqref{eq:conditional-pushforward}. The argument below supports independence of the learned coefficient and latent representations under the stated ratio condition; it does not establish that the second loss term performs exact marginal matching or that the learned conditional distribution equals the Bayesian posterior.

\begin{theorem}
\label{thm:independence-ratio}
Let $X$ and $Y$ have joint density $p_{XY}$ and marginal density $p_Y$. On points where the following density ratios are defined, $X$ and $Y$ are independent if and only if, for every $x$, $y$, and $y^{*}$,
\begin{equation}\label{eq:appendix-ratio}
\frac{p_{XY}(x,y)}{p_{XY}(x,y^{*})}
=
\frac{p_Y(y)}{p_Y(y^{*})}.
\end{equation}
\end{theorem}

\textit{Proof.}
If $X$ and $Y$ are independent, both sides of \cref{eq:appendix-ratio} equal $p_Y(y)/p_Y(y^{*})$. Conversely, suppose \cref{eq:appendix-ratio} holds. For fixed $x$,
\begin{equation}
\frac{p_{Y\mid X}(y\mid x)}{p_Y(y)}
=
\frac{p_{Y\mid X}(y^{*}\mid x)}{p_Y(y^{*})}
=:K(x).
\end{equation}
Hence $p_{Y\mid X}(y\mid x)=K(x)p_Y(y)$. Integrating with respect to $y$ gives
\begin{equation}
1=K(x)\int p_Y(y)\,\mathrm{d}y=K(x),
\end{equation}
so $p_{Y\mid X}(y\mid x)=p_Y(y)$ and $X$ and $Y$ are independent.
\hfill $\square$

Applied to the learned variables $(\boldsymbol{b},\boldsymbol{z})$, \cref{eq:appendix-ratio} yields the ratio condition in Eq.~\eqref{eq:ind-condition}. The term $\mathcal{L}_{\text{ind}_1}$ is designed to penalize violations of that condition. For each training pair $(\boldsymbol{m}_i,\boldsymbol{d}_i)$, the learned and projected coefficients are constructed from the same sample:
\begin{equation}
(\boldsymbol{b}_i,\boldsymbol{z}_i)=\mathcal{B}_{\theta}(\boldsymbol{m}_i),
\qquad
\boldsymbol{b}_i^{*}=\mathbf{T}_{\mathrm{o}}^{\dagger}\boldsymbol{d}_i.
\end{equation}
The projected coefficients define the empirical measure
\begin{equation}
\widehat{p}_{\mathrm{proj}}(\boldsymbol{b})
=
\frac{1}{N}\sum_{i=1}^{N}\delta_{\boldsymbol{b}_i^{*}}(\boldsymbol{b}).
\end{equation}
The second loss term in Eq.~\eqref{eq:ind-loss} is the paired empirical average
\begin{equation}
\mathcal{L}_{\text{ind}_2}
=
\frac{1}{N}\sum_{i=1}^{N}
\left[\log\frac{p_\theta(\boldsymbol{b}_i,\boldsymbol{z}_i)}
{p_\theta(\boldsymbol{b}_i^{*},\boldsymbol{z}_i)}\right]^2.
\end{equation}
It is used as an empirical optimization stabilizer. It is not a formal divergence between $p_\theta(\boldsymbol{b})$ and $\widehat{p}_{\mathrm{proj}}(\boldsymbol{b})$, and no exact marginal-matching guarantee is claimed.

If the learned representation satisfies
\begin{equation}
p_\theta(\boldsymbol{b},\boldsymbol{z})
=
p_\theta(\boldsymbol{b})p(\boldsymbol{z}),
\end{equation}
then conditioning on $\boldsymbol{b}=\boldsymbol{b}^{*}(\boldsymbol{d})$ leaves the latent reference distribution unchanged. Defining
\begin{equation}
g_{\theta,\boldsymbol{d}}(\boldsymbol{z})
=
\mathcal{B}_{\theta}^{-1}\!\left(\boldsymbol{b}^{*}(\boldsymbol{d}),\boldsymbol{z}\right),
\end{equation}
the GI-ION conditional distribution is
\begin{equation}
q_\theta(\cdot\mid\boldsymbol{d})
=
\left(g_{\theta,\boldsymbol{d}}\right)_{\#}p(\boldsymbol{z}).
\end{equation}
This pushforward is the learned conditional approximation used for sampling. Equality between $q_\theta(\cdot\mid\boldsymbol{d})$ and the Bayesian posterior $p(\boldsymbol{m}\mid\boldsymbol{d})$ would require additional approximation assumptions and is not asserted here.

\section{Maximum Mean Discrepancy (MMD)}
\label{app:MMD}

The MMD is a nonparametric statistic that quantifies the difference between two probability distributions based on samples drawn from each. Given finite samples $\{x_i\}_{i=1}^m \sim X$ and $\{y_j\}_{j=1}^n \sim Y$, a widely used estimator of the squared MMD is 

\begin{equation}
\begin{aligned}
 \mathrm{MMD}_b^2[X, Y] &= \frac{1}{m^2} \sum_{i, i'} k(x_i, x_{i'}) + \frac{1}{n^2} \sum_{j, j'} k(y_j, y_{j'}) \\ 
 &- \frac{2}{mn} \sum_{i, j} k(x_i, y_j).   
\end{aligned}
\end{equation}
where $k$ is the kernel function. In this study, the MMD loss is computed using the inverse multiquadratic (IMQ) kernel, defined as

\begin{equation}\label{eq: IMQ kernel}
k_{\mathrm{IMQ}}(x, y) = \frac{1}{\sqrt{\|x-y\|^2 + c^2}},
\end{equation}

\section{Metrics used in the paper}
\label{app:Metrics}

\subsection{Mean Squared Error}\label{app:mse}

 The MSE between the posterior mean \(\bar{\boldsymbol{m}}\) of generated samples  and the ground truth  \(\boldsymbol{m}^\ast\) is defined as

\begin{equation} \label{eq:mse}
\text{MSE} = \mathbb{E}\left[ \lVert \hat{\boldsymbol{m}} - \boldsymbol{m}^\ast \rVert_2^2 \right]
\end{equation}

\subsection{Continuous Ranked Probability Score}\label{app:CRPS}

The CRPS measures the accuracy of a model by comparing the predicted probability distribution with the actual observed value. Given the posterior samples $\{ \bm{m}\}_i^N$ and a ground truth \(\boldsymbol{m}^\ast\), the estimator is as follows:
\begin{equation}\label{eq:crps}
\text{CRPS} \approx \frac{1}{N} \sum_{i=1}^N \|\boldsymbol{m}_{i} - \boldsymbol{m}^\ast\| 
- \frac{1}{2N^2} \sum_{i=1}^N \sum_{j=1}^N \|\boldsymbol{m}_{i} - \boldsymbol{m}_{j}\|
\end{equation}

\subsection{Wasserstein Distance(Second-order)}\label{app:WD}
The second-order Wasserstein Distance measures the distance between the distribution of the branch network's output noise \( P \) and a standard Gaussian distribution \( Q \). Given the sample mean \( \hat\mu_P \) and covariance \( \hat\Sigma_P \) estimated from the output noise samples, the squared distance is calculated as:
\begin{equation}\label{eq:WD}
W_2^2(P, Q) = \| \hat\mu_P \|_2^2 + \operatorname{Tr}\Big( \hat\Sigma_P + I - 2 (\hat\Sigma_P)^{1/2} \Big).
\end{equation}

\subsection{Mutual Information}\label{app:MI}

Mutual information \( I(X;Y) \) quantifies the dependency between the outputs of the branch network \((\boldsymbol{b}_i, \boldsymbol{z}_i)\). It is estimated using the KSG1 estimator:
\begin{equation}\label{eq:MI}
\hat I^{\text{KSG1}} = \psi(k) - \frac{1}{N}\sum_{i=1}^N \Big[ \psi(n_x(i)+1) + \psi(n_y(i)+1) \Big] + \psi(N),
\end{equation}
where \(\psi\) is the digamma function, \(k\) is the number of nearest neighbors, and \(n_x(i), n_y(i)\) are neighbor counts in the marginal spaces.

\subsection{Structural Similarity Index}\label{app:SSIM}

The SSIM measures the structural similarity between the estimated model \(\hat{\boldsymbol{m}}\) and the ground truth \(\boldsymbol{m}^\ast\). It is defined as:
\begin{equation}\label{eq:SSIM}
\text{SSIM} = \frac{(2\mu_{\hat{\boldsymbol{m}}} \mu_{\boldsymbol{m}^\ast} + C_1)(2\sigma_{\hat{\boldsymbol{m}}\boldsymbol{m}^\ast} + C_2)}{(\mu_{\hat{\boldsymbol{m}}}^2 + \mu_{\boldsymbol{m}^\ast}^2 + C_1)(\sigma_{\hat{\boldsymbol{m}}}^2 + \sigma_{\boldsymbol{m}^\ast}^2 + C_2)},
\end{equation}
where \(\mu\) and \(\sigma^2\) represent the mean and variance, \(\sigma_{\hat{\boldsymbol{m}}\boldsymbol{m}^\ast}\) is the covariance, and \(C_1, C_2\) are stabilization constants.

\subsection{Signal-to-Noise Ratio}\label{app:SNR}

The SNR compares the power of the ground-truth signal \(\boldsymbol{m}^\ast\) to the power of the error in the estimated posterior mean \(\hat{\boldsymbol{m}}\):
\begin{equation}
\text{SNR} = 10 \log_{10} \left( \frac{\sum_{i=1}^N (\boldsymbol{m}^\ast_i)^2}{\sum_{i=1}^N ( \hat{\boldsymbol{m}}_i - \boldsymbol{m}^\ast_i )^2} \right).
\end{equation}

\section{PHYSICS-INFORMED LOSS OF EIKONAL EQUATION}
\label{app: pinn loss}
Here we present a systematic derivation of the physics-informed loss function for solving the Eikonal equation using GI-ION. The Eikonal equation governs the travel-time field $\bm{u}(\bm{x})$ of wavefronts in a medium with velocity field $\bm{m}(\bm{x})$:  

\begin{equation}\label{eq: Eikonal}
\left\{
\begin{array}{l}
|\nabla \bm{u}(\bm{x})| = \dfrac{1}{\bm{m}(\bm{x})}, \\[6pt]
\bm{u}(\bm{x}_0) = 0,
\end{array}
\right.
\end{equation}
where $\bm{x}^0$ denotes the source point. To avoid singularities and improve numerical stability, we square both sides:  

\begin{equation}\label{eq: Eikonal^2}
\left\{
\begin{array}{l}
\|\nabla \bm{u}(\bm{x})\|^2 = \frac{1}{\bm{m}^2(\bm{x})}, \\[6pt]
\|\bm{u}(\bm{x}_0)\|^2 = 0.
\end{array}
\right.
\end{equation}

For a given velocity model $\bm{m}_i$, the corresponding travel-time field $\bm{u}_i$ is obtained from GI-ION as follows:

\begin{equation}
\bm{u}_i = \mathcal{F}_{\bm{\theta}}(\bm{m}_i)(\bm{x})
= \sum_{j=1}^P b_j(\bm{m}_i) \cdot t_j(\bm{x}),
\end{equation}

The derivative of the k-th coordinate axis $x^k$ is:
\begin{equation}
\bm{u}_{x^k}^i(\bm{x}) = \frac{\partial \bm{u}_i}{\partial x^k} = \sum_{j=1}^P b_j(\boldsymbol{m}_i) \cdot  \frac{\partial t_j(\boldsymbol{x})}{\partial x^k}. 
\end{equation}

 In particular, for a 2-D problem with only two coordinate directions $x^1$ and $x^2$, so we can get the final form of the loss function:
\begin{equation}\label{eq: pinnloss}
\mathcal{L}_\text{physics} = \mathcal{L}_{\mathcal{N}} + \mathcal{L}_{\mathcal{B}}.
\end{equation}
where
\begin{equation}
\mathcal{L}_{\mathcal{N}} = \frac{1}{N} \sum_{i=1}^N \left( \left(\bm{u}_{x^1}^i(\boldsymbol{x}_\text{total})\right)^2 + \left(\bm{u}_{x^2}^i(\boldsymbol{x}_\text{total})\right)^2- \frac{1}{\bm{m}_i^2(\bm{x}_\text{total})}\right)^2. 
\end{equation}
\begin{equation}
\mathcal{L}_{\mathcal{B}} = \frac{1}{N}\sum_{i=1}^N \| \bm{u}_i(\bm{x}_\text{0}) \|^2.
\end{equation}

\clearpage
\section{Deep Generative Prior Network Architecture} \label{app: DGP network}

We implemented a VAE framework to construct a generator network that establishes mappings between low-dimensional latent spaces and high-dimensional signal spaces. Through this architecture, the trained model acquires the capability to (a) encapsulate statistical characteristics of the dataset, (b) synthesize sample signals with distributional fidelity to the original data, and (c) reformulate the ill-posed inverse problem as an optimization task seeking latent variables that produce generated signals optimally consistent with physical measurements. The encoder and decoder architectures used are shown in \cref{tab:encoder} and \cref{tab:decoder}.

\begin{table}[H]
\centering
\caption{Encoder Architecture}
\label{tab:encoder}
\resizebox{\textwidth}{!}{%
\begin{tabular}{|l|l|l|l|l|}
\hline
\textbf{Layer Name} & \textbf{Type} & \textbf{Input Shape} & \textbf{Output Shape} & \textbf{Activation} \\ \hline
enc\_conv1 & Conv2d & (1, 112, 112) & (32, 56, 56) & ReLU \\ \hline
enc\_conv2 & Conv2d & (32, 56, 56) & (64, 28, 28) & ReLU \\ \hline
enc\_conv3 & Conv2d & (64, 28, 28) & (128, 14, 14) & ReLU \\ \hline
enc\_conv4 & Conv2d & (128, 14, 14) & (256, 7, 7) & ReLU \\ \hline
enc\_fc1 & Linear & 256*7*7 & 1024 & ReLU \\ \hline
fc\_mu & Linear & 1024 & 20 & -- \\ \hline
fc\_logvar & Linear & 1024 & 20 & -- \\ \hline
\end{tabular}
}
\end{table}

\begin{table}[H]
\centering
\caption{Decoder Architecture}
\label{tab:decoder}
\resizebox{\textwidth}{!}{%
\begin{tabular}{|l|l|l|l|l|}
\hline
\textbf{Layer Name} & \textbf{Type} & \textbf{Input Shape} & \textbf{Output Shape} & \textbf{Activation} \\ \hline
dec\_fc1 & Linear & 20 & 1024 & ReLU \\ \hline
dec\_fc2 & Linear & 1024 & 256*7*7 & ReLU \\ \hline
dec\_conv1 & CT2d & (256, 7, 7) & (128, 14, 14) & ReLU \\ \hline
dec\_conv2 & CT2d & (128, 14, 14) & (64, 28, 28) & ReLU \\ \hline
dec\_conv3 & CT2d & (64, 28, 28) & (32, 56, 56) & ReLU \\ \hline
dec\_conv4\_1 & CT2d & (32, 56, 56) & (16, 112, 112) & ReLU \\ \hline
dec\_conv5\_1 & Conv2d & (16, 112, 112) & (8, 112, 112) & ReLU \\ \hline
dec\_conv6\_1 & Conv2d & (8, 112, 112) & (1, 112, 112) & Sigmoid \\ \hline
dec\_conv4\_2 & CT2d & (32, 56, 56) & (16, 112, 112) & ReLU \\ \hline
dec\_conv5\_2 & Conv2d & (16, 112, 112) & (8, 112, 112) & ReLU \\ \hline
dec\_conv6\_2 & Conv2d & (8, 112, 112) & (1, 112, 112) & Softplus \\ \hline
\end{tabular}
}
\end{table}

\clearpage
\section{Computational Efficiency and Scalability} \label{app: Computational Efficiency}

To evaluate the computational performance of GI-ION, we conducted a comparative analysis using the 2D random-field tomography experiment described in Section~\ref{sec:random-field-tomography}, which serves as the primary benchmark in this study. Table~\ref{tab:computation_2d} details the computational costs for GI-ION, the standard INN, and the MCMC baseline.

\begin{table}[H]
\centering
\caption{Computational Cost Comparison on 2D random-field Tomography}
\label{tab:computation_2d}
\resizebox{\textwidth}{!}{%
\begin{tabular}{@{}lccc@{}}
\toprule
\textbf{Method} & \textbf{Training Time} & \textbf{Inference Time$^*$} & \textbf{Total Cost$^\dagger$} \\ \midrule
GI-ION (Ours) & 28407s & 3.1327s & $\sim$25h \\
Standard INN & 7352s & 17.349s & $\sim$50h \\
MCMC & N/A & 986s & $\sim$2738h \\ \bottomrule
\multicolumn{4}{l}{\footnotesize $^*$ Average time to generate and retain 5000 posterior samples for a single observation.} \\
\multicolumn{4}{l}{\footnotesize $^\dagger$ Cost perspective for solving multiple inverse problems on the total test set.}
\end{tabular}
}
\end{table}

The training time of GI-ION ($\sim$ 7.9 hours) is higher than that of the standard INN ($\sim$ 2.0 hours), primarily due to the additional computation required for the physics-informed loss evaluation. However, both methods remain within the same order of magnitude, ensuring feasible offline training.

In a separate batch-validation setting, GI-ION processes 10,000 distinct observations, generating one inverse result per observation, in approximately 6 s, compared with approximately 35 s for the standard INN. These batch-validation timings include forward-network evaluation and velocity-field reconstruction; they refer to a different workload from the single-observation, 5,000-posterior-sample timings reported in Table~\ref{tab:computation_2d}. The lower batch-validation time of GI-ION is attributed to our specialized architecture and training strategy, which allow for a more compact network size without compromising accuracy. For the single-observation posterior-sampling setting reported in Table~\ref{tab:computation_2d}, MCMC requires approximately 16 minutes of continuous sampling. Furthermore, leveraging the parallel capabilities of neural networks, GI-ION can perform posterior sampling for multiple observations simultaneously, whereas MCMC is inherently sequential.

\clearpage
\section{Implementation Details}\label{app: implementation details}

To facilitate the reproducibility of our results, we provide a detailed summary of the hyperparameters, data preprocessing steps, and computational environment used in this study.
\Cref{tab:hyperparameters} summarizes the specific architecture configurations and optimization settings for the three experimental cases presented in Sections~\ref{sec:surface-wave}--\ref{sec:seismic-slice}. The weighting coefficients were empirically set to $\lambda_1 = 1$ (physics), $\lambda_2 = 10$ (data), and $\lambda_3 = 0.1$ (independence) for the 1D cases, and adjusted to $\lambda_3 = 0.5$ for the 2D case to account for the different dimensionality of the latent space.
All algorithms were implemented in Python using the \texttt{PyTorch} library. The numerical experiments were conducted on a Linux workstation equipped with Intel Xeon Gold 5320 CPU and a single NVIDIA GeForce RTX 3090 GPU.

\begin{table}[H]
\centering
\caption{Hyperparameters and Training Settings}
\label{tab:hyperparameters}
\resizebox{\textwidth}{!}{%
\begin{tabular}{@{}llll@{}}
\toprule
 & \shortstack{1D surface-wave\\dispersion} & \shortstack{2D random-field\\tomography} & \shortstack{2D seismic-slice\\tomography} \\
\midrule
\multicolumn{4}{@{}l}{\textit{GI-ION}} \\
\midrule
Trunk Network   & Identity       & FNN  & FNN\\
layers          & \textbackslash & 8    & 8   \\
units           & \textbackslash & 128  & 128 \\
Activation      & \textbackslash & ReLU & ReLU\\
Branch Network  & INN            & INN  & INN \\
layers          & 4              & 8    & 8   \\
Subnetwork(s, t)& FNN            & FNN  & FNN \\
units           & 128            & 128  & 256\\
Activation      & ReLU           & ReLU & ReLU \\
\midrule
\multicolumn{4}{@{}l}{\textit{INN}} \\
\midrule
layers          & 4 & 8    & 8   \\
units           & 128 & 512  & 512 \\
Activation      & ReLU & ReLU & ReLU\\
\midrule
\multicolumn{4}{@{}l}{\textit{Optimization}} \\
\midrule
Optimizer & Adam & Adam & Adam\\
Batch Size & 500 & 500 & 500 \\
Total Epochs & 1,000 & 3,000 & 3,000 \\
Init LR  & $10^{-3}$& $10^{-3}$& $10^{-3}$\\
Final LR & $10^{-5}$& $10^{-5}$& $10^{-5}$\\
\midrule
\multicolumn{4}{@{}l}{\textit{Data Splits}} \\
\midrule
Training Samples & $10^5$ / $20,000$ / $2,000$ & $10^5$ / $20,000$ / $2,000$ & $10^5$ \\
Testing Samples & 10,000 & 10,000 & 10,000 \\
\midrule
\multicolumn{4}{@{}l}{\textit{MCMC}} \\
\midrule
Iterations          & 120,000 & 220,000 & 220,000 \\
Burn-in iterations  & 20,000 & 120,000 & 120,000 \\
Thinning interval   & 20 & 20 & 20\\
\bottomrule
\end{tabular}
}
\textit{Note:}  All the INN layers take the RealNVP affine coupling blocks \cite{dinh2017density}.
\end{table}